\documentclass[graybox]{svmult}

\usepackage{mathptmx}       
\usepackage{helvet}         
\usepackage{courier}        
\usepackage{type1cm}        
\usepackage{makeidx}         
\usepackage{graphicx}        
\usepackage{multicol}        
\usepackage[bottom]{footmisc}
\usepackage{natbib}

\makeindex             

\usepackage[UKenglish]{babel}
\usepackage{amsmath}
\usepackage{latexsym}
\usepackage{amssymb}
\usepackage{graphicx}
\usepackage[all]{xy}
\usepackage{color}
\usepackage{latexsym}
\usepackage{amssymb}
\usepackage{amsmath}
\usepackage{color}
\usepackage{stmaryrd}

\usepackage[lined,boxed,commentsnumbered]{algorithm2e}
\usepackage{epic,eepic}
\newcommand{\PAL}{\textbf{PAL}}

\newcommand{\EL}{\textbf{EL}}
\newcommand{\ELKv}{\textbf{ELKv}}
\newcommand{\ELKvr}{\ensuremath{\ELKv^r}}

\newcommand{\MLKv}{\textbf{MLKv}}
\newcommand{\MLKvr}{\ensuremath{\MLKv^r}}
\newcommand{\MLKvrb}{\ensuremath{\MLKv^b}}

\newcommand{\EquiKw}{\ensuremath{{\texttt{Kw}\!\lra}}}

\newcommand{\PALKv}{\textbf{PALKv}}
\newcommand{\PALKvr}{\ensuremath{\textbf{PALKv}^r}}

\newcommand{\PILKvr}{\ensuremath{\textbf{PILKv}^r}}

\newcommand{\BP}{\textbf{P}}

\newcommand{\Ag}{\textbf{I}}

\newcommand{\SMLKvb}{\ensuremath{{\mathbb{SMLKV}^b}}}

\newcommand{\SELKvr}{\ensuremath{\mathbb{SELKV}^r}}

\newcommand{\SPLKw}{\ensuremath{\mathbb{SNCL}}}

\newcommand{\SPLKwT}{\ensuremath{\mathbb{SNCLT}}}
\newcommand{\SPLKwTr}{\ensuremath{\mathbb{SNCL}4}}

\newcommand{\SPLKwEuc}{\ensuremath{\mathbb{SNCL}5}}

\newcommand{\SPLKwTTr}{\ensuremath{\mathbb{SNCLS}4}}
\newcommand{\SPLKwTEuc}{\ensuremath{\mathbb{SNCLS}5}}

\newcommand{\Tr}{\texttt{wKw4}}
\newcommand{\Euc}{\texttt{wKw5}}
\newcommand{\KwB}{\texttt{KwB}}
\newcommand{\NECKvr}{${\texttt{NECKv}}^r$}

\newcommand{\EQUIV}{\ensuremath\texttt{SYM}}
\newcommand{\DISBK}{{\texttt{ATEUC}}}

\newcommand{\lr}[1]{\langle #1 \rangle}
\newcommand{\rel}[1]{\stackrel{#1}{\rightarrow}}

\newcommand{\simall}{\{\sim_i\mid i\in \Ag\}}

\newcommand{\mc}[1]{\mathcal{#1}}
\newcommand{\M}{\mc{M}}

\newcommand{\tr}[1]{\text{#1}}

\newcommand{\DISTK}{{\texttt{DISTK}}}

\newcommand{\TAUT}{{\texttt{TAUT}}}

\newcommand{\GENK}{{\texttt{NECK}}}
\newcommand{\RE}{{\texttt{RE}}}
\newcommand{\ATOM}{{\texttt{!ATOM}}}
\newcommand{\NEG}{{\texttt{!NEG}}}
\newcommand{\CON}{{\texttt{!CON}}}
\newcommand{\AK}{{\texttt{!K}}}

\newcommand{\SUB}{{\texttt{SUB}}}
\newcommand{\MP}{{\texttt{MP}}}

\newcommand{\AxT}{{\texttt{T}}}
\newcommand{\AxTr}{{\texttt{4}}}
\newcommand{\AxEuc}{{\texttt{5}}}

\newcommand{\AKvr}{\ensuremath{\texttt{!Kv}^r}}
\newcommand{\KwT}{\texttt{KwT}}
\newcommand{\KwTr}{\texttt{Kw4}}
\newcommand{\KwEuc}{\texttt{Kw5}}

\newcommand{\Kw}{\ensuremath{\Delta}}
\newcommand{\KwG}{\Kw}
\newcommand{\KKw}{\textit{Kw}}

\newcommand{\KwCon}{\texttt{KwCon}}
\newcommand{\KwDis}{\texttt{KwDis}}

\newcommand{\DISTKvr}{\ensuremath{\texttt{DISTKv}^r}}
\newcommand{\KvrTr}{\ensuremath{\texttt{Kv}^r4}}
\newcommand{\KvrDIS}{\ensuremath{\texttt{Kv}^r\lor}}
\newcommand{\KvrBot}{\ensuremath{\texttt{Kv}^r\bot}}

\newcommand{\NECU}{\ensuremath{\mathtt{NECU}}}

\newcommand{\POSTKh}{\ensuremath{\mathtt{POSTKh}}}

\renewcommand{\D}{{\textbf{C}}}

\newcommand{\N}{\mathcal{N}}
\newcommand{\Kv}{\ensuremath{\textit{Kv}}}

\newcommand{\K}{\ensuremath{\textit{K}}}
\newcommand{\hK}{\ensuremath{\hat{\textit{K}}}}

\newcommand{\EXIST}{\texttt{INC}}

\newcommand{\Dia}{\Diamond}
\newcommand{\C}{\textbf{C}}
\newcommand{\lra}{\ensuremath{\leftrightarrow}}

\newcommand{\bis}{\mathrel{\mathchoice%
{\raisebox{.3ex}{$\,
  \underline{\makebox[.7em]{$\leftrightarrow$}}\,$}}%
{\raisebox{.3ex}{$\,
  \underline{\makebox[.7em]{$\leftrightarrow$}}\,$}}%
{\raisebox{.2ex}{$\,
  \underline{\makebox[.5em]{\scriptsize$\leftrightarrow$}}\,$}}%
{\raisebox{.2ex}{$\,
  \underline{\makebox[.5em]{\scriptsize$\leftrightarrow$}}\,$}}}}

\newcommand{\Act}{\ensuremath{\mathbf{A}}}

\newcommand{\relall}{\{\to_i \mid i\in \Ag\}}
\newcommand{\relallp}{\{\to'_i \mid i\in \Ag\}}

\newcommand{\AxTransKU}{\ensuremath{\mathtt{4KU}}}
\newcommand{\AxEucKU}{\ensuremath{\mathtt{5KU}}}

\newcommand{\AxTrU}{\ensuremath{\mathtt{TU}}}

\renewcommand{\E}{{\mathcal E}}
\newcommand{\F}{{\mathcal F}}

\newcommand{\DISTU}{\ensuremath{\mathtt{DISTU}}}

\newcommand{\COMPKh}{\ensuremath{\mathtt{COMPKh}}}

\newcommand{\WSKh}{\ensuremath{\mathtt{WSKh}}}

\newcommand{\EMP}{\ensuremath{\mathtt{EMP}}}

\newcommand{\Kh}{\textit{Kh}}
\newcommand{\SKh}{\mathbb{SKH}}

\renewcommand{\S}{\mathcal{S}}
\renewcommand{\E}{\mathcal{E}}

\newcommand{\U}{\textit{U}}

\newcommand{\LKh}{\mathbf{L_{Kh}}}

\renewcommand{\phi}{\varphi}

\DeclareSymbolFont{symbolsC}{U}{txsyc}{m}{n}
\DeclareMathSymbol{\strictif}{\mathrel}{symbolsC}{74}

\newcommand{\NCL}{\textbf{NCL}}
\newcommand{\ML}{\textbf{ML}}

\newcommand{\Kd}{\ensuremath{\textit{Kd}}}

\newcommand{\bBox}[2]{{\Box_{#1}^#2}}
\newcommand{\bDia}[2]{{{\Diamond}_{#1}^#2}}

\begin{document}

\title*{Beyond knowing that:}
\subtitle{A new generation of epistemic logics\thanks{A mindmap of this paper is here: \url{http://www.phil.pku.edu.cn/personal/wangyj/mindmaps/}}}
\author{Yanjing Wang}
\institute{Yanjing Wang\at Department of Philosophy, Peking University, \email{y.wang@pku.edu.cn}}
%
%
\maketitle


\abstract{Epistemic logic has become a major field of philosophical logic ever since the groundbreaking work by \cite{Hintikka:kab}. Despite its various successful applications in theoretical computer science, AI, and  game theory, the technical development of the field  has been mainly focusing on the propositional part, i.e., the propositional modal logics of ``knowing that''. However, knowledge is expressed in everyday life by using various other locutions such as ``knowing whether'', ``knowing what'', ``knowing how'' and so on (knowing-wh hereafter). Such knowledge expressions are better captured in quantified epistemic logic, as was already discussed by \cite{Hintikka:kab} and his sequel works at length. This paper aims to draw the attention back again to such a fascinating but largely neglected topic. We first survey what Hintikka and others did in the literature of quantified epistemic logic, and then advocate a new quantifier-free approach to study the epistemic logics of knowing-wh, which we believe can balance expressivity and complexity,  and capture the essential reasoning patterns about knowing-wh. We survey our recent line of work on the epistemic logics of `knowing whether'', ``knowing what'' and ``knowing how'' to demonstrate the use of this new approach.}

\section{Introduction}
\label{sec:Intro}

Epistemic logic as a field was created and largely shaped by Jaakko Hintikka's groundbreaking work. Starting from the very beginning, \cite{Hintikka:kab} set the stage of epistemic logic in favor of a possible world semantics,\footnote{Hintikka was never happy with the term ``possible worlds'', since in his models there may be no ``worlds'' but only situations or states, which are partial descriptions of the worlds. However, in this paper we will still use the term ``possible worlds'' for convenience.} whose rich and intuitive structure facilitates in-depth philosophical discussions and an intuitive understanding of knowledge that leads to various applications in other fields such as distributed systems and artificial intelligence. In a nutshell, Hintikka's notion of knowledge amounts to the elimination of uncertainty. At a given world, the alternative relation induces a split of all the possible worlds: the epistemically possible ones and the rest. The agent knows $\phi$ at a world iff the $\neg \phi$ worlds are ruled out in its epistemic alternatives according to the agent. In fact, such a semantics also works for other propositional attitudes that are essentially about information, such as belief \citep{Hintikka2003}. 

\cite{Hintikka:kab} devoted most of the book on propositional epistemic logic with the following language (call it \EL):
$$\phi::=\top\mid p\mid \neg\phi\mid (\phi\land\phi)\mid K_i\phi$$
\noindent where $K_i \phi$ reads ``agent $i$ \textit{knows that} $\phi$''. The language is interpreted on Kripke models $\M=\langle S,\relall, V\rangle$ where $S$ is a non-empty set of possible worlds, $\to_i\subseteq S\times S$ and $V: \BP\to 2^S$. The semantics for $\K_i\phi$ is as follows:
$$
\begin{array}{|rcl|}
\hline
\mc{M},s\vDash K_i \phi &\Leftrightarrow&\text{for all } t \text{ such that }s\to_i t: \mc{M},t\vDash \phi  \\
\hline 
\end{array}
$$

According to \cite{Hintikka:kab}, $\to_i$ should be reflexive and transitive. In many applications, it is also reasonable to take it as an  equivalence relation, which gives rise to the $\mathbb{S}5$ axiom system, a very strong epistemic logic \citep{RAK}: 
\begin{center}
\begin{tabular}{lclc}
\multicolumn{4}{c}{System $\mathbb{S}5$}\\
\multicolumn{2}{l}{Axioms}&Rules&\\
\TAUT & \tr{ all the instances of tautologies}&\MP & $\dfrac{\phi, \phi\to\psi}{\psi}$\\
\DISTK & $K_i (p\to q)\to (K_ip\to K_iq)$&\GENK &$\dfrac{\phi}{ K_i\phi}$\\ 
\texttt{T} & $K_ip\to p$&\SUB &$\dfrac{\phi}{\phi [p\slash\psi]}$ \\
\texttt{4} &$K_ip\to K_iK_ip$&\phantom{$\dfrac{p}{\Box p}$}\\
\texttt{5} & $\neg K_ip\to K_i\neg K_ip$&\phantom{$\dfrac{p}{\Box p}$}\\
\end{tabular}
\end{center}

Despite various philosophical debates regarding the axioms \texttt{4} and \texttt{5}, and the problem of \textit{logical omniscience} (cf. \citep{Lenzen78}), propositional epistemic logic has been successfully applied to many other fields because its semantic notion of knowledge is intuitive and flexible enough to handle uncertainties in various contexts. The knowledge modality $\K_i$ is in particular powerful when combined with other modalities such as the temporal ones and the action modalities, which resulted in two influential approaches which can model changes of knowledge: \textit{Epistemic Temporal Logic} (ETL) and \textit{Dynamic Epistemic Logic} (DEL) (cf. e.g., \cite{RAK,DELbook}). See \cite{ELbook} for an overview of the applications of epistemic logic. 

\medskip

However, knowledge is not only expressed by ``knowing that''. For example, we often use the verb ``know'' with an embedded question such as: \\

\begin{minipage}{0.5\textwidth}
\begin{itemize}
\item I \textit{know whether} the claim is true. 
\item I \textit{know what} your password is. 
\item I \textit{know how} to swim. 
\end{itemize}
\end{minipage}
\begin{minipage}{0.5\textwidth}
\begin{itemize}
\item I \textit{know why} he was late. 
\item I \textit{know who} proved this theorem. 
\item I \textit{know where} she has been.
\end{itemize}
\end{minipage}
\medskip

In the rest of the paper, we call these constructions \textit{knowing-wh}: \textit{know} followed by a wh-question word.\footnote{\textit{How} is in general also considered as a wh-question word, besides \textit{what, when, where, who, whom, which, whose}, and \textit{why}.} The following table shows the number of hits returned by googling the corresponding terms.\footnote{The ``knows X'' search term can  exclude the phrases such as ``you know what'' and count only the statements, while ``know X'' may appear in questions as well.}
\begin{table}
\caption{Hits (in millions) returned by google}
\begin{tabular}{c@{\qquad}c@{\qquad}c@{\qquad}c@{\qquad}c@{\qquad}c@{\qquad}c}
\hline \noalign{\smallskip}
X & that & whether& what& how & who & why\\ 
\noalign{\smallskip}\svhline\noalign{\smallskip}
``know X'' & 574 & 28 & 592 & 490 & 112 & 113\\ 
``knows X'' & 50.7 & 0.51 & 61.4 & 86.3 & 8.48 & 3.55\\ 
\noalign{\smallskip}\hline\noalign{\smallskip}
\end{tabular}
\end{table}
From the statistics, at least ``know what'' and ``know how'' are equally frequent, if not more, as ``know that'' in natural language, and other expressions also play important roles in various contexts. Are those knowing-wh constructions as theoretically interesting as ``knowing that''? Below we will briefly look at it from three different perspectives of linguistics, philosophy, and AI. 

Linguists try to understand such constructions from a more general perspective in terms of classifications of verbs: which verbs can take an embedded wh-question? For example, \textit{forget, see, remember} are like \textit{know} in this sense. However, it is a striking cross-linguistic fact that the verb \textit{believe} cannot take any of those embedded questions, in contrast with philosophers' usual conception of knowledge in terms of strengthened justified true belief. Linguists have been trying to give explanations in terms of factivity and other properties of verbs with interesting exceptions (cf. e.g., \citep{Egre08} and references therein). Moreover, when \textit{know} is immediately followed by a noun phrase, it can usually be translated back to the knowing-wh constructions by treating the noun phrase as a \textit{concealed question}, e.g., knowing the price of milk can be treated as knowing \textit{what} the price of milk is \citep{heim1979}. The semantic variability of the same knowing-wh-construction in different contexts also interest linguists a lot, e.g., ``I know which card is the winning card'' can mean I know Ace is the winning card for the game, or I know the card that my opponent holds is the winning card. There are approaches that give uniform treatments to handle this kind of context-sensitivity (cf. e.g., \cite{Aloni01}). 

For philosophers, especially epistemologists, it is crucial to ask whether those knowing-wh statements are also talking about different kinds of knowledge. For example, it has been a frequently debated topic whether knowledge-how can be reduced to knowledge-that (cf. e.g.,  \citep{Ryle49,stanley2011know}). As another example, for philosophers of science, knowing why is extremely important, as it drives science forward. However, what amounts to knowing why? Many philosophers think knowing an \textit{scientific explanation} is the key to answering why-questions, and there is a large body of research on it  (cf. e.g., \citep{vanfraassen1980}). Knowing who also draws some attention from philosophers in analyzing the more general  propositional attitude ascriptions, see \citep{BL03}. 

Already in the early days of AI, researchers realized knowing-wh statements are useful in specifying the precondition or the effects of actions \citep{Moore77}. For example, it is crucial for a robot to know \textit{where} to check or \textit{whom} to ask, if it does not know \textit{what} the email address of the person it wants to contact. \cite{McCarthy79} even considered knowing what as the most important type of knowledge in AI. Such knowing-wh statements also show up in various implemented AI systems, e.g., knowledge-based planning system \citep{Petrick-Bacchus:2004,Petrick-Bacchus:2004b}. Besides constructing knowledge bases, it is very handy to specify the goal of a system using knowing-wh constructions, e.g., knowing whether is used quite frequently to specify knowledge goals and precondition for actions. 
\medskip

So, what about epistemic logicians? In fact, \cite{Hintikka:kab} devoted the last chapter to ``knowing who'' in the context of quantified epistemic logic, for the reason that the agent names are already in the epistemic language that he introduced earlier. Hintikka believed other knowing-wh constructions can be treated alike with different sorts of constants in place.\footnote{Later on he singled out ``knowing why'' in his framework of interrogative models \citep{Hintikka95}.} In fact he proposed to treat knowing-wh as ``one of the first problems'' in epistemic logic \citep{Hintikka1989}. The formalism involves quantifiers that \textit{quantify into} the modal scope which may cause ambiguity according to Quine.\footnote{See \citep{Garson2001} for a survey on the (technical) difficulties about quantification in modal logic.} Hintikka had lengthy discussions on conceptual and technical problems of quantified epistemic logic and in fact gradually developed a more general epistemic logic which he called a ``second generation epistemic logic'' \citep{Hintikka2003}. However, the quantified epistemic logic did not draw as much attention as its propositional brother. As a result, the classic textbook by \cite{RAK} has only a very brief discussion of first-order epistemic logic, and in the handbook of epistemic logic by \cite{ELbook}, there is not much about quantifiers either. The only dedicated survey that we found for quantified epistemic logic is a section in a long paper on epistemic logic by \cite{gochet2006epistemic}. It seems that the mainstream epistemic logicians mainly focus on the propositional cases. However, not only Hintikka himself did quite a lot of work on it but also there are fascinating new technical developments in quantified epistemic logic. This motivates the first part this paper: to give a brief overview on what Hintikka and others did about epistemic logics of knowing-wh and quantified epistemic logic in general. 

On the other hand, introducing quantifiers explicitly in the epistemic language has a high computational cost: many interesting quantified epistemic logics are not decidable. However, there is a way to go around this. In this paper we would like to propose a general quantifier-free approach to the logics of knowing-wh, which may balance expressivity and complexity. The central idea is simple: treat knowing-wh construction as new modalities, just like Hintikka did for knowing that. This approach can avoid some of the  technical and conceptual problems of the quantified epistemic logic due to its weak language. New techniques and logics are being developed as will be surveyed in the later part of the paper. 

\medskip 

The rest of the paper is organized as follows: In Section \ref{sec.related} we survey Hintikka's various works on knowing-wh.  Section \ref{sec.relatedother} reviews the recent technical developments of quantified epistemic logic. Section \ref{sec.ELwh} explains our new approach and its considerations. Section \ref{sec.example} gives three concrete examples to demonstrate our approach. In the last section we conclude with some further directions. 
\section{Background in quantified epistemic logic}
\subsection{Hintikka on knowing-wh}
\label{sec.related}
According to \cite{Hintikka1989}, one of the most important applications of epistemic logic is to understand questions.\footnote{It also makes sense to understand knowing-wh constructions by first understanding the semantics of questions, see \citep{Harrah02} and references therein. Knowing-wh is then knowing a\slash the answer of the corresponding wh-question.} A question ``Who is $b$?'' amounts to the request of information: bring about that I know who $b$ is. Hintikka called ``I know who $b$ is'' the \textit{desideratum} of the corresponding question. Under this view, the study of questions reduces largely to the study of their corresponding desiderata. This interest in the relationship between questions and knowledge also led Hintikka to the pursuit of a \textit{Socratic epistemology} that weighs \textit{knowledge acquisition} more importantly than \textit{knowledge justification} which has been the focus of the traditional epistemology \citep{hintikka2007socratic}. 

To formalize ``I know who $b$ is'' we do need quantifiers. \cite{Hintikka:kab} proposed the formula $\exists x \K (b=x)$, and compared it with $K\exists x (b=x)$ in order to demonstrate the distinction between \textit{de re} and \textit{de dicto} in the epistemic setting.\footnote{\cite{Hintikka:kab} argued that the quantification into the modal context is necessary and not misleading, in contrast to Quine who was against such quantification due to the lack of substitution of identity in modal context.} He called the earlier one \textit{knowledge of objects} and the later one propositional knowledge. However, once the constants and quantifiers are introduced into the language, we need a much richer structure over possible worlds. The possible worlds may not share the same domain of objects, for you may imagine something non-existent to exist in some possible world.\footnote{How the domain varies may affect the corresponding quantified modal axioms, see \citep{G07FML} for a overview on this issue in first-order modal logic.} Now how do we ``pick up'' an object in order to evaluate the formula ``$\exists x \K (b=x)$''? \cite{Hintikka1989} proposed to draw world-lines in different ways to identify objects across the world. His most important point here is that depending on how you draw the world-lines, the formulas like $\exists x \K(b=x)$ may have different meanings. For example, $\exists x \K (b=x)$ can mean I can visually identify a person, e.g., in a party scenario I can say I know who Bill is by pointing at someone: ``just that guy over there!''. According to \cite{Hintikka1989}, this requires to draw \textit{perspectival} world-lines to connect the visual images, which can sometimes be used to interpreted knowing who as \textit{acquaintance}. On the other hand, we can draw \textit{public} world-lines, which contribute to the semantics of knowing who by description, e.g., I know who Bill is: he is the mayor of this city and a well-know logician. We can also think that there are two kinds of quantifiers corresponding to these two ways of drawing the world-lines \citep{HinSym2003}. Since the formalism  of knowing-wh is still based on the knowing that operator $\K_i$, Hintikka did not consider them as a different type of knowing \citep{Hintikka2003}. 

Besides the simple knowing-wh sentences, there are natural knowing-wh expressions which involve predicates, e.g., ``I know who murdered $b$'' can be formalized as $\exists x K M(x, b)$, which is the desideratum of the question ``Who murdered $b$?''. To fulfill the desideratum $\exists x K M(x, b)$, is it enough to have $\K M(a,b)$ for some $a$? \cite{Hintikka1989b} argued that merely knowing that $M(a,b)$ does not always lead to knowing who: the questioner should also know who $a$ actually is, which is called the \textit{conclusiveness condition}. Indeed, answering the question ``Who gave the first speech?'' by ``The first speaker.'' may not be informative at all. Of course it is debatable whether this requirement is pervasive in most of the contexts. From this point of view, the existential generalization rule may not hold: $K M(a,b)$ does not entail $\exists x \K M(x, b)$. 

It becomes more interesting when complicated knowing-wh sentences are considered. An example given by  \cite{Hintikka2003} is ``I know whom every young mother should trust'' (with the intention pointing to ``her own mother''). It seems that we need to pick up the trusted one in a uniform way for each young mother, and thus $\exists f K\forall x (M(x)\to T(x,f(x)))$ is a faithful formalization. Actually such knowledge of functions is pervasive in empirical sciences, where the research can be viewed as asking Nature what is the (functional) dependence between different variables \citep{Hintikka1999}. For example, let $x$ be the controlled variable and $y$ be the observed variable, and we ask Nature the dependence between $x$ and $y$ by doing experiments $E$ by changing the value of $x$. The desideratum of such a question is that I know the dependence between $x$ and $y$ according to the experiments, which can be formalized as $\exists f \K\forall x E(x, f(x))$ where $E$ can be viewed as the relation paring the values of $x$ and $y$ according to the experiments. Like before, merely having $\K\forall x E(x, g(x))$ is not enough, we do need a conclusiveness condition that you know the function $g$: $\exists f \K \forall x(f(x)=g(x)).$ In this way, \cite{Hintikka95:KAK} managed to explain how mathematical knowledge, such as the knowledge of certain functions, plays a role in empirical research.  

However, the above discussion leads to the introduction of higher-order entities, whose existence is unclear \citep{Hintikka2003}. To avoid this problem, Hintikka made use of the idea prominent in the \textit{Independence Friendly Logic} proposed by \cite{hintikkaSandu1989}. The idea is to introduce the \textit{independence sign} ``$\slash$'' into the logic language to let some quantifiers jump out of the scopes of earlier ones, in order to have a branching structure of quantifiers which are linearly ordered in the formulas. For example, $\forall x  (\exists y\slash {\forall x})(x=y)$ is not valid anymore,  compared to $\forall x \exists y(x=y)$, since the choice of $y$ is \textit{independent} from the choice of $x$. Now the earlier ``young mother'' formula becomes $\K\forall x (\exists y\slash\K)(M(x)\to T(x,y))$ without the second-order quantification. Likewise, the desideratum of an experiment can be formalized as $\K(\forall x) (\exists y\slash \K)E(x,y)$. The slash sign not only works with quantifiers but also logical connectives. For example, $\K( p(\lor\slash\K)\neg p)$ expresses knowing whether $p$ while $\K( p\lor \neg p)$ amounts to knowing a tautology. There is also a beautiful correspondence between the desideratum and the presupposition of the same wh-question. The desideratum can usually be obtained by adding a suitable slash in the corresponding presupposition. For example, the presupposition of ``Who murdered $b$?'' is that $\K \exists x M(x, b)$, i.e., I know someone murdered $b$, and the desideratum is $\K (\exists x\slash K) M(x, b)$， which is equivalent to $\exists x \K M(x, b)$, i.e., I know who murdered $b$. \cite{Hintikka2003} called the epistemic logic using such an extended language the \textit{second generation epistemic logic}, for it can go beyond the first-order epistemic logic, although the apparent quantifications are still first-order.\footnote{The above $\K\forall x (\exists y\slash\K)(M(x)\to T(x,y))$ is an example that cannot be expressed in standard first-order epistemic logic. }

\subsection{Recent technical advances of quantified epistemic logic}
\label{sec.relatedother}
The only comprehensive survey on quantified epistemic logic that we found is the section 5 of a paper by \cite{gochet2006epistemic}, which covers most of the important works up to the beginning of this century.\footnote{For the background of first-order modal logic, the readers are referred to the handbook chapter by \cite{G07FML} and the book by \cite{FittingM1998}. For the discussions on the philosophical issues of quantified first-order epistemic logic, see \cite{HollidayP2014} and references therein.} Here we supplement it with some of the recent advances, which are, however,  by no means exhaustive.

Most of the recent developments in quantified epistemic logic are application-driven. To handle cryptographic reasoning, \cite{CoDa07} propose a complete first-order epistemic logic with a counterpart semantics in order to model the indistinguishability of messages modulo one's decoding ability. To formalize the reasoning in games, \cite{KanekoN96} propose a first-order epistemic logic with common knowledge. \cite{Wolter00} shows that even very simple fragments of such a FO epistemic logic are not decidable. On the other hand, decidable fragments are found using techniques by \cite{SturmWZ00} based on the idea of \textit{monodic fragments} of quantified modal logic, where only one free variable is allowed to appear in the scope of modalities. In a similar way, some monodic fragments of first-order temporal logic are proved decidable (cf. e.g., \citep{HodkinsonWZ00,HodkinsonWZ02,Hodkinson02a}). It also inspired \cite{BelardinelliL11} to discover useful fragments of FO epistemic temporal logic. FO epistemic temporal logic has also been used to verify security properties as demonstrated by \cite{BelardinelliL09,BelardinelliL12}.  

In propositional epistemic logic, agent names are like rigid designators and they actually are indexes of the epistemic alternative relations in the model. However, this limits epistemic logic to a fixed, finite set of agents. Moreover, agents cannot have uncertainty about each other's identity. A natural extension is to allow (implicit) quantification over agents \citep{Corsi02,CorsiO13,CorsiT14}, where different readings of a quantified modal formula can also be disambiguated. Another quantifying-over-agent approach appears in the context of rough sets with multiple sources (as agents) by \cite{Khan2010}. 

It is also interesting to quantify over propositions,  which leads to second-order epistemic logic by \cite{Belardinelli:2015,BelardinelliH16}, built on an early work by \cite{Fine1970}. In such a framework, one can express that currently $i$ knows everything that $j$ knows,\footnote{Modeling it globally can be done in propositional modal logic with new axioms like $K_jp\to K_ip$, cf. e.g., \cite{LomuscioR99}.} which was handled earlier in a different approach by \cite{vanDitmarsch2012}. 

Recent years also witness the growth of \textit{inquisitive semantics} as an interdisciplinary field between linguistics and logic. It gives a uniform semantics to both descriptive and interrogative sentences (cf. e.g., \citep{CiardelliGR13}). In such a framework, one can combine knowing that operator with an embedded interrogative compositionally, and this is how knowing whether is treated in the epistemic inquisitive logic  \citep{Ciardelli14,CiardelliR15}. The readers are referred to the PhD thesis by \cite{Ivano16} for recent developments.  

\section{Epistemic logics of knowing-wh: a new  proposal}
\label{sec.ELwh}
Our point of departure from the aforementioned existing research is that we take a knowing-wh construction as a \textit{single} modality, just like $\K$ for knowing that, without explicitly introducing quantifiers, predicates, and equality symbols into the logic language. For example, instead of rendering ``agent $i$ knows what the value of $c$ is'' as $\exists x K_i(c=x)$, we simply have $Kv_i c$ where $\Kv_i$ is a new knowing what modality. An example language of knowing what is as follows (to be discussed in detail  later): 
$$\phi::=\top\mid p\mid \neg\phi\mid (\phi\land\phi)\mid \K_i\phi\mid \Kv_i c$$
\noindent where $c$ belongs to a set $\C$ of constant symbols.  

Following Hintikka, we take a semantics-driven approach for there is usually not enough syntactic intuition on the possible axioms for such knowing-wh constructions. We can discover interesting axioms by axiomatizing the valid formuals w.r.t.\ the semantics. The models are usually richer than Kripke models for propositional epistemic logic. For example, the semantics for $\Kv_i c$ is given over first-order Kripke models with a constant domain: 
$\M=\langle S,D,\relall, V, V_C\rangle$ where $\langle S,\relall, V\rangle$ is a usual Kripke model, $D$ is a constant domain of values (all the worlds share the same $D$), and $V_\C:\C\times S\to D$ assigns to each (non-rigid designator) $c\in \C$ a $d\in D$ on each $s\in S$: 
$$
\begin{array}{|lll|}
\hline
\M,s\vDash Kv_i c &\Leftrightarrow& \text{for any}~t_1,t_2: ~\text{if}~s\to_it_1,s\to_it_2, \text{then}~V_C(c,t_1)=V_C(c,t_2).\\
\hline
\end{array}
$$
Intuitively, $i$ knows what the value of $c$ is iff $c$ has the same value over all the $i$-accessible worlds. This is the same as the semantics for $\exists x K_i (c=x)$ on constant domain FO Kripke models. We will come back to the details in Section \ref{sec.kwhat}. 

After defining the language and semantics, we can try to find a complete axiomatization with meaningful axioms, and then dynamify the logic to include updates of knowledge as in dynamic epistemic logic \citep{DELbook}. The axioms will tell us some intrinsic logical features of the knowing-wh construction. We may come back to philosophy with new insights after finishing the formal work. 
\medskip

Such an approach has the following advantages: 
\begin{description}
\item[\textbf{Neat language and characterizing axioms}] Using knowing-wh modalities can make the formal languages very simple yet natural, which can also highlight the logical differences between different knowing-wh in terms of intuitive axioms, e.g., knowing whether $\phi$ is equivalent to knowing whether $\neg \phi$. It will also become clear how knowing-wh modalities differ from the normal modalities, e.g., knowing how to achieve $\phi$ and knowing how to achieve $\psi$ does not entail knowing how to achieve $\phi\land\psi$ (e.g., take $\psi=\neg\phi$). 
\item[\textbf{Balancing expressivity and complexity}] The new languages may be considered as small fragments of quantified epistemic logic and we can try to balance the expressivity and complexity. For example, the above \Kv\ modality packages a quantifier, a $\K$ modality, and an equality together. Such a packed treatment is also the secret of the success of standard modal logic, where a quantifier and a relational guard are packed in a modality. Such weaker languages are in general more applicable in practice due to computational advantages. Our approach may also help to discover new decidable fragments of quantified modal logics. 
\item[\textbf{Avoiding some conceptual problems}] The history of epistemic logic taught us a lesson that the logical framework can be extremely useful even before philosophers reach a consensus on all its  issues, if they ever do so at all. Certain conceptual difficulties about quantified epistemic logic  should not stop us from developing the logic further while bearing those questions in mind, since new insights may come as you start to move forward. Our weaker languages are free of explicit quantifiers, thus it may avoid some difficulties in the full quantified epistemic logic and makes us focus on the limited but reasonably clear fragments.\footnote{The absence of equality symbols also make the substitution of equal constants apparently irrelevant.}
\item[\textbf{Connections to existing modal logics}] As we will see, each of the knowing-wh logics has some very close (sometimes surprising) friends in propositional modal logic. We may benefit from the vast existing results and tool support for propositional modal logic. As we will see, the new operators can also motivate new ways to update the models which were not considered before. 
\end{description}

Of course, there are also limitations and difficulties of this approach: 
\begin{itemize}
\item The languages cannot express knowing-wh constructions in a fully compositional way when complicated constructions are involved, e.g., John knows what Mary knows about logic. Also from the linguistic perspective, our approach cannot handle context-sensitivity of the meaning of the knowing-wh constructions.\footnote{See \citep{Aloni16} in this volume for a quantified epistemic logic treatment of this context-sensitivity of knowing who, using conceptual covers proposed by \cite{Aloni01}.}
\item Our languages are relatively weak, but the models are very rich in order to accommodate an intuitive semantics. This apparent asymmetry between syntax and semantics may cause difficulties in axiomatizating the logics. However, we may restore the symmetry by simplifying the models modulo the same logic once we have a complete axiomatization w.r.t.\ the rich models. 
\item From a syntactic point of view, the new logics are usually not normal, e.g., knowing whether $\phi\to \psi$ and knowing whether $\phi$ does not entail knowing whether $\psi$, for you may know that $\phi$ is false but have no idea about the truth value of $\psi$.\footnote{\label{foot}On the other hand, a slightly different axiom holds intuitively: knowing whether $\phi\lra \psi$ and knowing whether $\phi$ does entail knowing whether $\psi$.}
\item Although many knowing-wh modalities share a general form of $\exists x \K \phi(x)$, different modalities can still behave quite differently depending on the exact shape of $\phi(x)$. Also, the existential quantifier may not necessarily be first-order as in the later example of a logic of knowing how. 
\item In some cases it is highly non-trivial to give a reasonable semantics since we do not understand enough about the meaning of certain knowing-wh yet.
\end{itemize}
In the following we give three example studies on knowing-wh to demonstrate the claimed advantages, and how we overcome some of the technical difficulties mentioned above. 


\section{Examples}
\label{sec.example}
In this section, we demonstrate the use and techniques of the proposed approach with three examples: the logics of knowing whether, knowing what, and knowing how. Besides the historical background and the common pursuit for complete axiomatizations, each example has its own special focus to give the readers a more general picture of the approach. The readers may pay attention to the points below.
\begin{itemize}
\item Knowing whether:  expressivity comparisons over models and frames w.r.t.\ standard modal logic, and  completeness proof for such non-normal modal logic; 
\item Knowing what: interaction axioms between knowing that  and the new modality, conditionalization of the new modality, asymmetry between syntax and semantics and the techniques to restore the symmetry, and a new update operation;
\item Knowing how: philosophically inspired language design, AI inspired semantics design, epistemic models without epistemic relations, and techniques of completeness proof when $x$ is not unique (nor first-order) to make $\exists x \K \phi(x)$ true. 
\end{itemize}
Impatient readers who only want to see one example may jump  to Section \ref{sec.kwhat} on a logic of knowing what since it is the most representative one for the proposed approach. In the following examples, we will focus on the ideas behind definitions and results rather than technical details, which can be found in the cited papers.   
\subsection{Knowing whether}
The logic of knowing-whether is perhaps the closest knowing-wh friend of the standard epistemic logic, yet it can already demonstrate many shared features of the logics of knowing-wh. Although it is clear that knowing whether $\phi$ ($\KKw_i\phi$) is equivalent to knowing that $\phi$ or knowing that $\neg\phi$, introducing the knowing whether operator firstly has an advantage in succinctness,  as \cite{DFHI14} showed. In many epistemic puzzles such as muddy children, the goal and the preconditions of actions are often formulated as knowing whether formulas.  As a philosophical example, \cite{vDHI} showed that although it is not possible to know every true proposition according to Fitch's paradox based on Moore sentences,\footnote{Fitch proved that you cannot know all the truths, e.g., $p\land\neg K_i p$ is not knowable by $i$, which is demonstrated by the inconsistent Moore sentence: $K_i(p\land\neg K_i p)$ in the basic epistemic logic, see \citep{fitch1963a}.} everything is eventually knowable in terms of knowing whether it is true (the truth value may have changed). It also makes sense to iterate the knowing whether operators of  different agents to succinctly capture the higher-order observability of agents towards each other, e.g., I know whether you know whether $p$ although I do not know whether $p$ (cf. e.g., the \textit{sees} operator by \cite{HerzigLM15}).
As a technical example, \cite{HHS96:knowingWT} made use of the alternations of knowing whether operators to neatly build $2^{\aleph_0}$ many mutually inconsistent knowledge states of two agents, which greatly simplified a previous construction by \cite{Aumann89} using knowing that operators. The construction of \cite{HHS96:knowingWT} relies on an intuitive axiom about knowing whether: $Kw_i\phi\lra Kw_i\neg\phi$. Now, what is the complete axiomatic system for the logic of knowing whether, where $\KKw_i$ is the \textit{only} primitive modality? How is the expressivity of this logic compared to that of the standard epistemic \slash modal logic? 


Actually, such technical questions have been partly addressed under the name of \textit{non-contingency logic} where the modality symbol $\Delta$ takes the place of \KKw, which we will follow from now on. Indeed, if you view the modal operator $\Box$ as a necessity operator  then $\Delta\phi:=\Box\phi\lor\Box\neg\phi$ says that $\phi$ is not \textit{contingent}. In different contexts this  operator has different readings. In the context of alethic modality, the study of contingency logic goes back to \cite{MR66} and involves the works of many well-known logicians;\footnote{For example, \cite{cresswell1988necessity,Kuhn95,Humberstone95,Demri97,Zolin99,Pizzi2007}, see \citep{FWvD15} for a survey.} in the epistemic context, it amounts to knowing whether,\footnote{See \cite{AloniEJ13} for more general versions of the knowing whether operator.} and its negation amounts to a notion of ignorance \citep{hoeketal:2004}; in the doxastic setting, $\Kw \phi$ says that the agent is opinionated about $\phi$; in the deontic setting, $\neg \Kw \phi$ means moral indifference \citep{Wright51}; in the proof theoretical context, $\neg \Kw \phi$ means that $\phi$ is undecided \citep{Zolin01}. In different settings, different frame conditions may be imposed, thus it is interesting to see how the logic behaves over different frame classes, as in standard modal logic. In the following, let us get a taste of this simple yet interesting language by looking into a few formal results.


\subsubsection{Language, semantics and expressivity}
Following the tradition in non-contingency logic, call the following language $\NCL$:
$$\phi::=\top\mid p\mid\neg\phi\mid(\phi\wedge\phi)\mid\KwG_i\phi$$
where $p\in\BP$ and $i\in\Ag$. It is interpreted on Kripke models $\M=\lr{S, \{\to_i\mid i\in\Ag\}, V}$:\footnote{Similar semantics has been applied to neighborhood structures \citep{FanD15}.}     
$$\begin{array}{|lcl|}
\hline
\mc{M},s\vDash\KwG_i\phi&\Leftrightarrow&\text{for all }t_1,t_2\text{ such that }s\to_it_1,s\to_it_2: (\mc{M},t_1\vDash\phi\Leftrightarrow\mc{M},t_2\vDash\phi)\\   
&\Leftrightarrow& \text{ either } \text{for all }t\text{ s.t. }s\to_it: \mc{M},t\vDash\phi \text{ or for all }t\text{ s.t. }s\to_it: \mc{M},t\nvDash\phi\\
\hline
\end{array}
$$
Note that we do not impose any properties on the frames unless specified. 
$\NCL$ is clearly no more expressive than the standard modal logic (\ML) since we can define a translation $t: \NCL\to \ML$ such that: $t(\Kw_i\phi)=\Box_it(\phi)\vee\Box_i\neg t(\phi)$. What about the other way around? If we restrict ourselves to reflexive models, we can also define a translation $t':  \ML \to \NCL$, namely  $t'(\Box_i\phi)=t'(\phi)\land \Kw_it'(\phi)$. However, \NCL\ and \ML\ do not have the same expressive power over arbitrary models. We can use a notion of bisimulation to measure the expressive power of the logic. Let us first recall the standard definition of bisimulation in modal logic: 
\begin{definition}[Bisimulation] Let $\M=\lr{S,\relall,V}$, $\N=\lr{S', \relallp, V'}$ be two models.
A binary relation $Z$ over $S\times S'$ is a \emph{bisimulation} between $\M$ and $\N$, if $Z$ is non-empty and whenever $sZs'$:
\begin{itemize}
\item (Invariance) $s$ and $s'$ satisfy the same propositional variables;
\item (Zig) if $s\to_it$, then there is a $t'$ such that $s'\to_it'$
and $tZt'$;
\item (Zag) if $s'\to_it'$, then there is a $t$ such that $s\to_it$
and $tZt'$.
\end{itemize}
$\M,s$ is \textit{bisimilar} to $\N,t$ ($\M,s\bis\N,t$) if there is a bisimulation between $\M$ and $\N$ linking $s$ with $t$. 
\end{definition}
It is well-known that modal logic is invariant under bisimilarity, thus bisimilarity is also an invariance relation for \NCL. However, it is too strong even on finite models. The two pointed models $\M, s$ and $\N, s'$ below satisfy the same \NCL\ formulas but they are not bisimilar.\footnote{Note that if there is at most one successor of $s$ then every $\Kw\phi$ formula holds.}
$$
\xymatrix{
\underline{s}:p\ar[r]|i& t:p & & \underline{s'}:p
}
$$
 However, in most of the cases when there are two or  more successors standard bisimilarity works fine. To tell the subtle difference we need to connect $\Delta$ with $\Box$. \cite{FWvD14} have a crucial observation that $\Box_i$ is \textit{almost definable} by $\Kw_i$. 
\begin{proposition}[Almost-definability Schema (AD) \cite{FWvD14}]
For any $\phi,\psi$ in the modal language with both $\Box_i$ and $\Delta_i$ modalities: $$\vDash\neg \Kw_i\psi\to(\Box_i \phi\lra(\Kw_i\phi\land \Kw_i(\psi\to\phi))).$$
\end{proposition}
The idea is that \textit{if} there are two $i$-accessible worlds differentiated by a formula $\psi$, then $\Box_i$ is locally definable in terms of $\Kw_i$. The missing part between $\Box_i\phi$ and $\Kw_i\phi$ is that we need to force $\phi$, instead of $\neg \phi$, to hold over the $i$-accessible worlds, and the contingency of $\psi$ helps to fill in the gap. This almost-definability schema (AD) inspires us to find: 
\begin{itemize}
\item a notion of $\KwG_i$-\textit{bisimulation} which characterizes the expressive power of \NCL; 
\item the suitable definition of canonical relations in the completeness proofs;
\item the right axioms for special frame properties. 
\end{itemize}

From AD, if there are two states which can be told apart by a \NCL\ formula then the standard bisimulation should work locally.  However, to turn  this precondition into a purely \textit{structural} requirement is quite non-trivial. The idea is to define the bisimulation notion within a \textit{single} model and then generalize the bisimilarity notion using disjoint unions of two  models.   

\begin{definition}[$\Delta$-Bisimulation]\label{Kw-bis} Let $\M=\lr{S,\{\to_i\mid i\in\Ag\},V}$ be a model.
A binary relation $Z$ over $S$ is a \emph{$\Delta$-bisimulation} on $\M$, if $Z$ is
non-empty and whenever $sZs'$:
\begin{itemize}
\item (Invariance) $s$ and $s'$ satisfy the same propositional variables;
\item (Zig) if \textit{there are two different successors $t_1,t_2$ of $s$ such that $(t_1,t_2)\notin Z$ and} $s\to_it$, then there exists $t'$ such that $s'\to_it'$ and $tZt'$;
\item (Zag) if \textit{there are two different successors $t'_1, t'_2$ of $s'$ such that $(t'_1,t'_2)\notin Z$ and} $s'\to_it'$, then there exists $t$ such that $s\to_it$ and $tZt'$.
\end{itemize}
$\M,s$ and $\N,t$ are $\Delta$-bisimilar ($\M,s\bis_\Delta\N,t$) if there is a $\Delta$-bisimulation on the \textit{disjoint union} of $\M$ and $\N$ linking $s$ and $t$.   
\end{definition}
In contrast to the standard bisimilarity, to show that $\Delta$-bisimilarity is indeed an equivalence relation is not at all trivial but a good exercise to appreciate better the definition.\footnote{The transitivity is hard, you need to enrich the two bisimulations a bit in connection with the middle model when proving it, see \citep{FanPhD}.}
  
Based on $\Delta$-bisimilarity, \cite{FWvD14} proved: 
\begin{theorem}
For image-finite (or \NCL\ saturated models) $\M, s$ and $\N, t$: $\M,s \bis_\Delta\N,t\iff \M,s\equiv_{\NCL}\N,t$ (satisfying the same \NCL\ formulas).
\end{theorem}   
\begin{theorem}
\NCL\ is the $\Delta$-bisimilarity invariant fragment of \ML\ (and \textbf{FOL}). 
\end{theorem}
The proof mimics the standard proofs in modal logic by using AD repeatedly to simulate $\Box$ whenever  possible. 


A natural question arises: if you can almost always define $\Box$ using $\Kw$ locally on models, is the difference in expressivity just a negligible subtlety? However, \cite{FWvD14} showed that in terms of frame definability it is a significant difference.  
\begin{theorem}
The frame properties of seriality, reflexivity, transitivity, symmetry, and Euclidicity are \textbf{not} definable in $\NCL$.
\end{theorem}
The proof by \cite{FWvD14} uses the following frames: 
$$
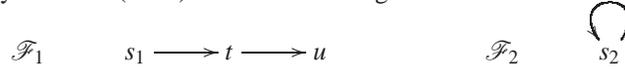
\xymatrix{
\F_1&s_1\ar[r]& t\ar[r] &u && \F_2&s_2 \ar@(ur,ul)
}$$
It can be shown that $\F_1\vDash\phi\iff \F_2\vDash\phi$ for all \NCL-formula $\phi$, based on the invariance under $\Delta$-bisimilarity and possible valuations over the frames. However, the left frame is not reflexive (transitive, serial, symmetric and Euclidean) while the right one has all these properties. Therefore such frame properties are not definable. This presents a sharp difference between \NCL\ and \ML, and this may cause difficulties in axiomatizing $\NCL$ over different frame classes.  
\subsubsection{Axiomatizations}
In axiomatizing $\NCL$ over different frame classes to apply it in different contexts, we apparently face the following difficulties: 
\begin{itemize}
\item It is impossible to use \NCL\ formulas to capture frame properties. 
\item \NCL\ is not normal, e.g.,  $\KwG_i(\phi\to\psi)\land \KwG_i\phi\to \KwG_i\psi$ is invalid, as mentioned before. 
\item \NCL\ is also not strictly weaker than modal logic, i.e. $\KwG_i\phi\lra\KwG_i\neg\phi$ is valid.
\end{itemize}
The following system \SPLKw\ is proposed by \cite{FWvD14,FWvD15}\footnote{See \cite{FWvD15} for comparisons with other equivalent systems in the literature.} 


\begin{center}
\begin{tabular}{lc@{\qquad}lc}
\multicolumn{4}{c}{System $\SPLKw$}\\
\multicolumn{2}{l}{Axioms}&Rules&\\
\TAUT & \tr{ all the instances of tautologies}&\MP & $\dfrac{\phi, \phi\to\psi}{\psi}$\\
\KwCon & $\KwG_i(q\to p)\land\KwG_i(\neg q \to p)\to\KwG_i p$ &\texttt{NEC} &$\dfrac{\phi}{ \Kw_i\phi}$\\ 
\KwDis & $\KwG_ip\to \KwG_i (p\to q )\lor \KwG_i(\neg p\to q)$&\SUB &$\dfrac{\phi}{\phi [p\slash\psi]}$ \\
\EquiKw & $\KwG_ip\lra\KwG_i\neg p$& \texttt{REPL}&$\dfrac{\phi\lra\psi}{\KwG_i\phi\lra\KwG_i\psi}$\\
\end{tabular}
\end{center}
\KwCon\ tells us how to derive $\Kw_i\phi$, and \KwDis\ tells us how to derive from $\Kw_i\phi$. $\KwCon$ is actually useful if we take it as a guide for the questioning strategy aiming at knowing whether $p$ (cf. e.g., \citep{LiuW13}). Imagine that a student $i$ wants to know whether he has passed the exam ($p$) or not, but does not want to ask the teacher directly. According to the axiom, he can ask the teacher two apparently innocent questions related to whether someone else (say $j$) has passed the exam ($q$): (1) ``Is it the case that $j$ or I passed the exam?'' (to obtain $\Kw_i (q\lor p)$, i.e., $\Kw_i(\neg q\to p)$) and (2) ``Is it the case that if $j$ passes then I pass too?'' (to obtain $\Kw_i (q\to p)$). By axiom $\KwCon$, $\Kw_ip$ then holds.\footnote{Here we can also see the parallel of deduction and interrogation that  \cite{hintikka2007socratic} discussed.} Note that since the distribution axiom no longer holds for $\Kw_i$, we need the replacement rule \texttt{REPL} to facilitate the substitution of equivalent formulas. 
  \begin{theorem}[\cite{FWvD15}]
\SPLKw\ is sound and strongly complete w.r.t. \NCL\ over the class of arbitrary frames. 
\end{theorem}
The completeness proof is based on the following canonical model construction, inspired by the almost-definability schema again.
\begin{definition}[Canonical model]\label{def.testdef} Define $\M^c=\lr{S^c,R^c,V^c}$ as follows:
\begin{itemize}
\item $S^c=\{s\mid s \text{ is a maximal consistent set of }\SPLKw\}$  
\item For all $s,t\in S^c$, $sR_i^c t$ iff \textit{there exists $\chi$} such that:
\begin{itemize}
\item $\neg\KwG_i \chi\in s$, and   
\item for all $\phi$, $\KwG_i \phi\land\KwG_i (\chi\to\phi)\in s$ implies $\phi\in t$.   
\end{itemize}
\item $V^c(p)=\{s\in S^c\mid p\in s\}$.
\end{itemize}
\end{definition}   
Readers who are familiar with modal logic can immediately see the similarity to the standard definition of canonical relations: $\Kw_i\phi\land \Kw_i(\chi\to \phi)$ acts as $\Box_i\phi$ given $\neg \Kw_i\chi\in s$. Note that if $\Kw_i\chi\in s$ for every \NCL-formula $\chi$ then there is simply no need to have an outgoing arrow from $s$.  

In the proof of the truth lemma, the hard part is to show that $\Delta_i\psi\not\in s$ implies $\M^c,s\nvDash\Delta_i\psi$. Here it is worthwhile to stress a characteristic feature which is shared by some other knowing-wh logics. Note that to show $\M^c,s\nvDash\Delta_i\psi$ (existence lemma), we need to construct \textit{two} successors of $s$ such that $\psi$ holds on one and does not hold on the other. Bearing the schema AD in mind, it boils down to show the following two sets are consistent, which can be proved using the axioms: 
\begin{enumerate}
\item \label{truthlem.one} $\{\phi\mid\Delta_i\phi\land\Delta_i(\psi\to\phi)\in s\}\cup\{\psi\}$ is consistent.
\item \label{truthlem.two} $\{\phi\mid\Delta_i\phi\land\Delta_i(\neg\psi\to\phi)\in s\}\cup\{\neg\psi\}$ is consistent.
\end{enumerate}   



For \NCL\ over other frame classes, \cite{FWvD15} present all the complete axiomatizations based on \SPLKw\ in Table \ref{tb.ax}.\footnote{For some equivalent proof systems in the literature, see the survey and comparisons in \citep{FWvD15}.} 
{\small
\begin{table}
\caption{Axiomatizations of \NCL\ over various frame classes}
\label{tb.ax}
$\begin{array}{|lll|l|}
  \hline
  \text{Notation}& \text{Axiom Schemas}& \text{Systems} & \text{Frames} \\
\hline
  \KwT & \KwG_i\phi\land\KwG_i(\phi\to\psi)\land\phi\to\KwG_i\psi& \SPLKwT=\SPLKw+\KwT & \text{reflexive}\\
  \KwTr & \KwG_i\phi\to\KwG_i(\KwG_i\phi\vee\psi)& \SPLKwTr=\SPLKw+\KwTr & \text{transitive}\\
  \KwEuc & \neg\KwG_i\phi\to\KwG_i(\neg\KwG_i\phi\vee\psi)&\SPLKwEuc=\SPLKw+\KwEuc & \text{euclidean}\\
   \Tr  & \KwG_i\phi\to\KwG_i\KwG_i\phi &\SPLKwTTr=\SPLKw+\KwT+\Tr & \text{ref. \& trans.}\\
   \Euc   & \neg\KwG_i\phi\to\KwG_i\neg\KwG_i\phi &\SPLKwTEuc=\SPLKw+\KwT+\Euc & \text{equivalence} \\
  \KwB  & \phi\to\KwG_i((\KwG_i\phi\land\KwG_i(\phi\to\psi)&\mathbb{SNCLB}=\SPLKw+\KwB & \text{symmetric}\\
     &\land\neg\KwG_i\psi)\to\chi)&&\\
  \hline
\end{array}
$
\end{table}
}
Note that although $\Tr$ and $\Euc$ look like the corresponding axioms $\texttt{4}$ and $\texttt{5}$ of standard epistemic logic, $\SPLKw+\Tr$ and $\SPLKw+\Euc$ are \textit{not} complete over the classes of transitive and euclidean frames respectively. We need their stronger versions. On the other hand, in presence of \KwT, $\Tr$ and $\Euc$ are enough to capture \NCL\ over $\mathbb{S}5$ frames. 

Here are two points we want to stress (details can be found in \citep{FWvD15}): 
\begin{itemize}
\item We may find new axioms by using the almost-definability schema to translate the standard modal logic axioms corresponding to the frame properties. 
\item The axioms are usually not canonical but we can  transform the canonical model into the right shape. 
\end{itemize}
 

We conclude our discussion on knowing whether by adding public announcements to \NCL: 
$$\phi::=\top\mid p\mid\neg\phi\mid(\phi\wedge\phi)\mid\KwG_i\phi\mid [\phi]\phi$$
with the standard semantics as in public announcement logic of \cite{Plaza89:lopc}:
$$
\begin{array}{|rcl|}
\hline
\mc{M},s\vDash [\psi]\phi &\Leftrightarrow& \M,s\vDash\psi  \textrm{ implies } \mc{M}|_\psi,s\vDash \phi\\
\hline
\end{array}
$$
\noindent where $\mc{M}|_\psi=(S',\{\to'_i\mid i\in \Ag\}, V')$ such that:
$S'=\{s\mid \M,s\vDash\psi\}$, $\to'_i=\to_i|_{S'\times S'}$ and $V'(p)=V(p)\cap S'$.

With the usual reduction axioms and the following one,  \cite{FWvD15} axiomatized the extended logic over various classes of frames: 
$$[\phi]\KwG_i\psi\lra(\phi\to(\KwG_i[\phi]\psi\vee\KwG_i[\phi]\neg\psi))$$
A similar story holds if we introduce the event model  modality in DEL \citep{FWvD15}. By having both the updates and knowing whether modalities in place, this simple language can be used to model the goal and the preconditions of actions in the scenarios of epistemic planning with polar questions/binary tests. For example, in a version of muddy children, the father asks ``Please step forward, if you \textit{know whether} you are dirty''. After repeating the announcement several times, all the dirty children \textit{know whether} they are dirty. 

Instead of the standard announcement operator, we can also introduce the \textit{announcing whether} operator $[?\phi]$ which updates the model with the $\phi$ or $\neg\phi$ depending on the actual truth value of $\phi$  (cf. e.g., \citep{van2010situation,vDF16}). It is easy to see that $[?\phi]\psi\lra ([\phi]\psi\land [\neg \phi]\psi)$. This operator may be useful in presenting protocols involving telling the truth value of a proposition such as the protocol for dining cryptographers \citep{DC}. In the next section, we will generalize this idea to announcing the value of a constant.

\subsection{Knowing what}
\label{sec.kwhat}
Knowing whether $\phi$ can also be viewed as  knowing \textit{what} the truth value of $\phi$ is. In this subsection, we survey the line of work on a simple yet ubiquitous type of knowing what: ``knowing [what the] value [is]'' where each constant has a value that ranges over a possibly infinite domain.\footnote{As we mentioned earlier, knowing the value can be seen as knowing the answer to a concealed question, see \cite{aloniinterpreting} and references therein for some recent discussions.} Note that since the domain may be infinite, it does not make sense to encode knowing the value of $c$ by the disjunction of knowing that $c=v_1$, knowing that $c=v_2$, and so on. This is a fundamental difference between knowing whether and knowing value, which makes the latter much more interesting.  

The study of knowing value as a modal operator dates back to \citep{Plaza89:lopc}, which is well-known for the invention of public announcement logic (\PAL).\footnote{A similar definition of the knowing value modal operator appeared earlier in \cite{XiwenW83} as an abbreviation in a setting of quantified epistemic logic.} Interestingly enough, almost one half of this classic paper was devoted not to ``knowing that'' but to ``knowing value'', which was, to our knowledge, largely neglected by the later literature except the  comments by  \cite{Dit07:comments}. \cite{Plaza89:lopc} used two running examples to demonstrate the update effects of public announcements: the muddy children and the sum-and-product puzzle.\footnote{Two people S and P are told respectively the sum and product of two natural numbers which are known to be below 100. The following conversation happens: P says: ``I do not know the numbers." S says: ``I knew you didn't." P says: ``I now know the numbers." S says: ``I now also know it."} To model the second puzzle, \cite{Plaza89:lopc} introduced a special $Kv_i$ modality to the epistemic language to express that agent $i$ knows the value of some constant. Let us call the following language $\PALKv$ (where $c$ is any constant symbol in a given set $\C$):
$$\phi::=\top\mid p\mid \neg\phi\mid (\phi\land\phi)\mid \K_i\phi\mid \Kv_i c\mid  [\phi]\phi$$
\noindent We use the usual abbreviations $\hK_i$ and $\lr{\phi}$ for the diamond versions of $\K_i$ and $[\phi]$.

By having both $\K_i$ and $\Kv_i$, $\PALKv$ can express interesting interactions between them, e.g., ``$i$ knows that $j$ knows the password but $i$ doesn't know what exactly it is'' by $\K_i\Kv_jc \land \neg\Kv_ic$. Note that replacing $\Kv$ by $\K$ and replacing constant $c$ by a proposition $p$ will result in an inconsistent formula $\K_iK_jp\land \neg K_ip$.\footnote{On the other hand, replacing $\Kv$ with the knowing whether operator results in a consistent formula. } 

In contrast to the in-depth study of public announcement logic, Plaza did not give the axiomatization of the above logic with both announcement and the $Kv$ operator but only a few axioms on top of $\mathbb{S}5$, and this was the starting point of the study by \cite{WF13}. It turns out that those axioms are not enough to capture the logic w.r.t.\ the semantics we mentioned at the beginning of Section \ref{sec.example} for $\Kv_i$: 
\begin{theorem}[\cite{WF13}]\label{thm.incomp}
The valid formula $\lr{p}\Kv_i c\land\lr{q}\Kv_i c\to\lr{p\lor q}\Kv_i c$ is not derivable in the $\mathbb{S}5$ system with Plaza's new axioms.
\end{theorem}

By defining a suitable bisimulation notion,  \cite{WF13} showed that \PALKv\ is not reducible to its announcement-free fragment \ELKv, thus the standard reductive-technique of dynamic epistemic logic cannot work here: you can never use reduction axioms to capture the logic of \PALKv\ based on a system of the epistemic logic with $\Kv_i$ but not announcements.\footnote{\cite{Plaza89:lopc} gave the following two extra introspection axioms on top of $\mathbb{S}5$ to capture this announcement-free fragment without a proof: $\Kv_ic\to \K_i\Kv_i c$ and $\neg \Kv_i c\to \K_i\neg\Kv_i c$. Our later language will supersede this simple language.} In the following, we propose an apparently more general conditional $\Kv_i$ operator that can encode the public announcements with reduction axioms. We believe the generalized operators constitute a language which is easier to use. 


\subsubsection{Language, semantics and expressivity}
We start with a conditional generalization of $\Kv_i$ operator introduced by \cite{WF13} (call the language \ELKvr\ where $^r$ means ``relativized''): 
$$\phi::=\top\mid p\mid \neg\phi\mid (\phi\wedge\phi)\mid \K_i\phi\mid \Kv_i(\phi,c)$$
\noindent where $\Kv_i(\phi,c)$ says ``agent $i$ knows what $c$ is \textit{given} $\phi$''. For example, I may forget my login password for a website, but I can still say that I know what the password is given that it is four-digit, since I have only one four-digit password ever. Actually, everyday knowledge is usually conditional.\footnote{For example, I know that I have hands given that I am not a brain in a vat.} 
As mentioned earlier, the semantics is based on first-order epistemic models with a constant domain $\M=\langle S,D,\simall, V, V_C\rangle$ where $\sim_i$ is an equivalence relation:
$$
\begin{array}{|lll|}
\hline
\M,s\vDash \Kv_i(\phi,c)&\Leftrightarrow&\text{for any}~ t_1,t_2\in S \text{ such that }s\sim_it_1\text{ and }s\sim_it_2:\\&&\M, t_1\vDash\phi \text{ and } \M,t_2\vDash\phi\text{ implies } V_C(c, t_1)=V_C(c,t_2)\\
\hline
\end{array}
$$
Intuitively, the semantics says that $i$ knows the value of $c$ given $\phi$ iff on all the $\phi$-worlds that he considers possible, $c$ has exactly the same value. The announcement operator can also be added to \ELKvr\ and obtain \PALKvr: 
$$\phi::=\top\mid p\mid \neg\phi\mid (\phi\wedge\phi)\mid \K_i\phi\mid \Kv_i(\phi,c)\mid [\phi]\phi$$
\PALKvr\ looks more expressive than \PALKv, but in fact both logics are equally expressive as the announcement-free \ELKvr: 
\begin{theorem}[\cite{WF13}]
The comparison of the expressive power of those logics are summarized in the following (transitive) diagram:
\begin{center}
$
\begin{array}{lll}
\ELKvr & \longleftrightarrow& \PALKvr\\
\uparrow       &                    & \updownarrow\\
\ELKv   & \longrightarrow    & \PALKv\\
\end{array}
$
\end{center}
\end{theorem}
It means that we can forget about \PALKv\ and use \ELKvr\ instead, qua expressivity. 

\subsubsection{Axiomatization}
An axiomatization for the multi-agent \ELKvr\ is given by \cite{WF14}: 

\begin{minipage}{0.6\textwidth}
\begin{center}
\begin{tabular}{lc}
\multicolumn{2}{c}{System $\SELKvr$-$\mathbb{S}5$}\\
\multicolumn{2}{l}{Axiom Schemas}\\
\texttt{TAUT} & \tr{ all the instances of tautologies}\\
\DISTK &$ \K_i (p\to q)\to (\K_i p\to \K_i q$)\\
\AxT&$\K_ip\to p$ \\
\AxTr&$\K_ip\to \K_i\K_ip$ \\
\AxEuc&$\neg \K_ip\to \K_i\neg \K_ip$ \\
\DISTKvr& $\K_i (p\to q)\to (\Kv_i(q,c)\to \Kv_i(p,c))$\\
\KvrTr& $\Kv_i(p,c)\to \K_i \Kv_i(p,c)$\\
\KvrBot& $\Kv_i(\bot,c)$\\
\KvrDIS& $\hK_i(p\wedge q)\wedge \Kv_i(p,c)\wedge \Kv_i(q,c)\to \Kv_i(p\vee q,c)$\\
\end{tabular}
\end{center}
\end{minipage}
\hfill
\begin{minipage}{0.3\textwidth}
\begin{center}
\begin{tabular}{lc}
\multicolumn{2}{l}{Rules}\\
\texttt{MP}& $\dfrac{\phi, \phi\to \psi}{\psi}$\\
\GENK &$\dfrac{\phi}{\K_i \phi}$\\
\SUB &$\dfrac{\phi}{\phi [p\slash\psi]}$\\
\RE &$\dfrac{\psi\lra\chi}{\phi\lra \phi[\psi\slash\chi]}$
\end{tabular}
\end{center}
\end{minipage}
\medskip

\noindent where \DISTKvr\ is the distribution axiom for the conditional $\Kv_i$ operator, which capture the interaction between $\K_i$ and $\Kv_i$ (note the positions of $p$ and $q$ in the consequent). $\KvrTr$ is a variation of the positive introspection axiom, and the corresponding negative introspection is derivable. \KvrBot\ stipulates that the $\Kv_i$ operator is essentially a conditional. Maybe the most interesting axiom is $\KvrDIS$ which handles the composition of the conditions: suppose all the epistemically possible $p$-worlds agree on what $c$ is and all the epistemically possible $q$-worlds also agree on $c$, then the overlap between $p$-possibilities and $q$-possibilities implies that all the $p\lor q$-possibilities also agree on what $c$ is. The careful reader may spot similarity between this axiom and the formula to show incompleteness in Theorem  \ref{thm.incomp}.

\cite{WF14} then showed the completeness of the above system: 
\begin{theorem}
$\SELKvr$ is sound and strongly complete for $\ELKvr$. 
\end{theorem} 
The highly non-trivial proof of the above theorem demonstrates the asymmetry between the syntax and semantics that we mentioned earlier. First note that in the canonical model, merely maximal consistent sets cannot work. The following is a model where \textit{two} logically equivalent states are needed to falsify $\Kv_1c$, where $c$ is assigned value $\circ$ and $\bullet$ respectively. This can never be embedded into a canonical model where states are maximal consistent sets. This problem is due to the fact that our language is too weak to capture all the information in the models.    
$$\xymatrix@C+5pt{
{p, c\mapsto\circ}\ar@(ur,ul)|{1,2}\ar@{-}[r]|1&{p, c\mapsto \bullet}\ar@(ur,ul)|{1,2}
}$$
The proof idea comes when we realize what those $\Kv_i(\phi, c)$ formulas actually are. Here, the perspective of quantified epistemic logic helps. Essentially, $\Kv_i(\phi,c)$ can be viewed as $\exists x \K_i(\phi\to c=x)$ where $x$ is a variable and $c$ is a \textit{non-rigid} constant. The $\Kv_i$ operator packages a quantifier, a modality, an implication and an equality together, without allowing the subformulas to appear freely. To build a suitable canonical model, we need to saturate each maximal consistent set with some extra information which roughly correspond to some subformulas of $\exists x \K_i(\phi\to c=x)$:
\begin{itemize}
\item counterparts of atomic formulas such as $c=x$;
\item counterparts of $\K_i(\phi\to c=x)$.
\end{itemize}
Moreover, we need to make sure these extra pieces of information are ``consistent'' with the maximal consistent sets and the canonical relations, by imposing further conditions. \cite{WF14} introduced two functions $f$ and $g$ to tell the current value of each $c$, and the potential value of $c$ given $\phi$ according to $i$. Thus a state in the canonical model is a triple $\lr{\Gamma, f, g}$ where $f$ and $g$ function as subformulas $c=x$ and $K_i(\phi\to  c=x)$. The extra conditions need to impose the consistency between such ``subformulas'' and the corresponding maximal consistent sets, e.g., $\psi\land\Kv_i(\psi, c) \in\Gamma$ implies $f(c)=g(i,\psi,c)$: if $\psi$ holds on the current world, then the value of $c$ given $\psi$ should be the same as the current value of $c$. 

Then we can prove the following statements:
\begin{itemize}
\item Each maximal consistent set can be properly saturated with some $f$ and $g$. 
\item Each saturated MCS including $\neg K_i\neg\phi$ has a saturated $\phi$-successor.
\item Each saturated MCS including $\neg Kv_i(\phi,c)$ has \textit{two} saturated $\phi$-successors which disagree on the value of $c$.
\end{itemize}
As in the case of knowing whether, the last ``existence lemma'' requires us to build two successors simultaneously based on some consistent sets, where axiom $\KvrDIS:\hK_i(p\wedge q)\wedge \Kv_i(p,c)\wedge \Kv_i(q,c)\to \Kv_i(p\vee q,c)$ plays an important role. See \cite{WF14} for details. 

\medskip

Coming back to the original question by Plaza, we can now axiomatize multi-agent \PALKvr\ by adding the following reduction axiom schemas easily:\footnote{Uniform substitution does not work for these new schemas.}
\begin{center}
\begin{tabular}{lc}
\ATOM & $\lr{\psi}p\lra(\psi\land p)$\\
\NEG & $\lr{\psi}\neg \phi\lra(\psi\land \neg \lr{\psi}\phi)$\\
\CON & $\lr{\psi}(\phi\land\chi)\lra(\lr{\psi}\phi\land \lr{\psi}\chi)$\\
\AK & $\lr{\psi}\K_i \phi\lra (\psi\land \K_i( \psi\to\lr{\psi}\phi))$\\
\AKvr& $\lr{\phi}\Kv_i(\psi,c)\lra (\phi\land \Kv_i(\lr{\phi}\psi, c))$
\end{tabular}\\
\end{center}       
Note that the specific values do not show in the language,  and this gives us the hope to build models with a small domain and a small set of possible worlds for each satisfiable \ELKvr\ formulas. It can be shown that $\ELKvr$ is not only decidable but with a complexity not higher than standard modal logic.\footnote{The decidability of \ELKvr\ over epistemic models was shown by \cite{Xiong14}.} 
\begin{theorem}[\cite{Ding15}]
\ELKvr\ over arbitrary models is \textsc{Pspace}-complete. 
\end{theorem}
\subsubsection{Simplification of the semantics}
As we mentioned, the models for \ELKvr\ are rich, but the language is quite weak, thus some information in the model cannot be expressed. To restore the symmetry between semantics and syntax, we may try to simplify the models while keeping the same logic intact (valid formulas). As we will see, the simplified semantics may sharpen our understanding of the logic and facilitate further technical discussions. 
\medskip

Let us start with a simple but crucial observation that we already touched  implicitly in the discussion of the completeness proof: $\neg \Kv_i(\phi, c)$ can be viewed as a special diamond formula, since it says that there are two $i$-accessible $\phi$-worlds that do not agree on the value of $c$.\footnote{In some applications in computer science, the exact value is also not that important, but people care about whether two values are \textit{equivalent}, e.g., see logic works on data words \citep{Bojanczyk:2011,Bojańczyk2013}. The author thanks Martin Otto for pointing this out.} Note that the semantics does not really rely on the exact value of $c$ on each world, but it does depend on whether $c$ has the \textit{same} value. This inspires \cite{GW16} to propose a simplified semantics, which interprets the corresponding diamond $\bDia{i}{c}$ w.r.t.\ a ternary relation $R_{i}^c$ in the Kripke models, where  $sR_{i}^cuv$ intuitively means that $u$, $v$ are two $i$-successors of $s$, which do not agree on the value of $c$.\footnote{Instead of the ternary relation, it seems also natural to introduce an anti-equivalence relation $R^c$ such that $sR^ct$ intuitively means that $s$ and $t$ do not agree on the value of $c$. However, this approach  faces troubles due to the limited expressive power of the modal language, see \citep{GW16} for a detailed discussion. } 

Let us consider the following language \MLKvr (essentially a disguised rewritten version of \ELKvr\ by replacing $\K_i$ with $\Box_i$, and $\neg Kv_i$ with $\Diamond^c_i$)
$$\phi::=\top\mid p\mid \neg\phi\mid (\phi\wedge\phi)\mid \Box_i\phi\mid \bDia{i}{c}\phi$$

The models are propositional Kripke models with both binary and ternary relations $\langle S,\{\to_i:i\in \Ag\},\{R_{i}^{c}:i\in \Ag,c\in\D\},V\rangle$, where $\to_i$ is as before for the $\Box_i$ operator. To simplify discussions, we do not assume $\to_i$ to be an equivalence relation in this subsection. The semantics for $\bDia{i}{c}\phi$ is as follows:  
$$
\begin{array}{|lll|}
\hline
\M,s\vDash\bDia{i}{c}\phi& \iff& \exists u,v: \text{ such that }  sR_{i}^{c}uv, \M,u\vDash\phi \text{ and } \M,v\vDash\phi\\
\hline
\end{array}
$$
To maintain the same logic (valid formulas modulo the rewriting), the following three conditions on $R_i^c$ are imposed. 
\begin{enumerate}
\item Symmetry: $sR_i^c vu$ iff $sR_i^c uv$; 
\item Inclusion: $sR_{i}^{c}uv$ only if $s\to_i u$ and $s\to_i v$; 
\item Anti-Euclidean property: $sR_{i}^{c}t_1t_2$ and $s\to_i u$ implies that at least one of $sR_{i}^{c}ut_1$ and  $sR_{i} ^{c}ut_2$ holds.
\end{enumerate}

The first two conditions are intuitive, given the intention of $R^c_i$. The condition (3) is the most interesting one and it is depicted as follows: 
$$
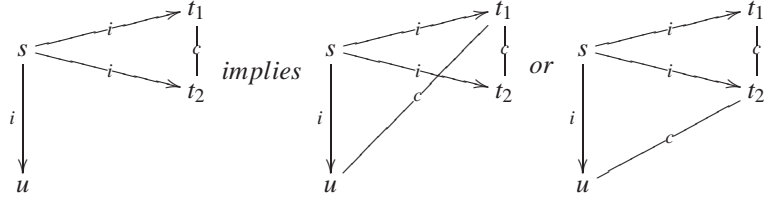
\xymatrix@R-20pt@C+30pt{
&t_1\ar@{-}[dd]|c\\
s \ar[ru]|i \ar[rd]|i\ar[dddd]_i&\\    
&t_2\\
&\\
&\\
u
}
\xymatrix@R-20pt@C+30pt{
\\
\\    
implies\\
\\
\\
}
\xymatrix@R-20pt@C+30pt{
&t_1\ar@{-}[dd]|c\ar@{-}[dddddl]|c\\
s \ar[ru]|i \ar[rd]|i\ar[dddd]_i&\\    
&t_2\\
&\\
&\\
u
}
\xymatrix@R-20pt@C+30pt{
\\
\\    
or \\
\\
\\
}\xymatrix@R-20pt@C+30pt{
&t_1\ar@{-}[dd]|c\\
s \ar[ru]|i \ar[rd]|i\ar[dddd]_i&\\    
&t_2\ar@{-}[dddl]|c\\
&\\
&\\
u
}
$$
It says that if two $i$-accessible worlds do not agree on the value of $c$ then any third $i$-accessible world must disagree with one of the two worlds on $c$.

Given a first-order Kripke model for \ELKvr, we have a corresponding Kripke model with both binary and ternary relations for \MLKvr, by defining $R^c_{i}$ as $\{ (s,u,v)\mid s\to_i u, s\to_i v, \text{ and } V_\C(c, u)\not=V_\C(c, v)\}$. Such a computed relation satisfy the above three properties.\footnote{Clearly the corresponding models also satisfy more properties, such as $sR_{i}^{c}uv$ only if $v\not=u$. However, (1)-(3) are enough to keep the logic intact, see \citep{GW16} for details.} Moreover, \cite{GW16} show that the following proof system (essentially the translated version of S5-free \SELKvr) is sound and strongly complete w.r.t.\ the Kripke models satisfying $(1)-(3)$. 

\begin{minipage}{0.64\textwidth}
\begin{center}
\begin{tabular}{lc}
\multicolumn{2}{c}{Another look of \SELKvr}\\
\multicolumn{2}{l}{Axiom Schemas}\\
\texttt{TAUT} & \tr{ all the instances of tautologies}\\
\DISTK &$ \Box_i (p\to q)\to (\Box_i p\to \Box_i q)$\\
$\DISTKvr$ &$ \Box_i(p\to q)\to (\bDia{i}{c} p \to \bDia{i}{c}q$)\\
\KvrBot& $\neg \bDia{i}{c} \bot $\\
$\KvrDIS$ &$\Diamond_i (p\land q)\land \bDia{i}{c}(p\lor q)\to  (\bDia{i}{c}p\lor \bDia{i}{c}q)$\\
\end{tabular}
\end{center}
\end{minipage}
\hfill
\begin{minipage}{0.23\textwidth}
\begin{center}
\begin{tabular}{lc}
\multicolumn{2}{l}{Rules}\\
\texttt{MP}& $\dfrac{\phi, \phi\to \psi}{\psi}$\\
\GENK &$\dfrac{\phi}{\Box_i \phi}$\\
\SUB &$\dfrac{\phi}{\phi [p\slash\psi]}$\\
\RE &$\dfrac{\psi\lra\chi}{\phi\lra \phi[\psi\slash\chi]}$
\end{tabular}
\end{center}
\end{minipage}
\\

We can massage the system into an equivalent form to make it look more familiar by adding the necessitation rule \NECKvr, deleting the \KvrBot, and changing the shape of $\DISTKvr$ (see \citep{GW16} for the proof of equivalence):\\

\begin{minipage}{0.64\textwidth}
\begin{center}
\begin{tabular}{lc}
\multicolumn{2}{c}{Massaged \SELKvr}\\
\multicolumn{2}{l}{Axiom Schemas}\\
\texttt{TAUT} & \tr{ all the instances of tautologies}\\
\DISTK &$ \Box_i (p\to q)\to (\Box_i p\to \Box_i q)$\\
$\DISTKvr$ &$ \Box_i(p\to q)\to (\bBox{i}{c} p\to \bBox{i}{c}q$)\\
$\KvrDIS$ &$\Dia_i (p\land q)\land \bDia{i}{c}(p\lor q)\to  (\bDia{i}{c}p\lor \bDia{i}{c}q)$\\
\end{tabular}
\end{center}
\end{minipage}
\hfill
\begin{minipage}{0.30\textwidth}
\begin{center}
\begin{tabular}{lc}
\multicolumn{2}{l}{Rules}\\
\texttt{MP}& $\dfrac{\phi,\phi\to\psi}{\psi}$\\
\GENK &$\dfrac{\phi}{\Box_i \phi}$\\
\NECKvr &$\dfrac{\phi}{\bBox{i}{c}\phi}$\\
\SUB &$\dfrac{\phi}{\phi [p\slash\psi]}$\\
\RE &$\dfrac{\psi\lra\chi}{\phi\lra \phi[\psi\slash\chi]}$\\
\end{tabular}
\end{center}
\end{minipage}


\medskip

It seems that $\bBox{i}{c}$ (the dual of $\bDia{i}{c}$) almost behaves just like a normal modality. However, the distribution axiom $\bBox{i}{c}(p\to q)\to (\bBox{i}{c} p\to \bBox{i}{c}q)$ is not valid. This is because $\bDia{i}{c}$ is essentially a \textit{binary} diamond, but we force the two arguments to be the same! To restore the normality, we can consider the following language: (\MLKvrb): 
$$\phi::=\top\mid p\mid \neg\phi\mid (\phi\wedge\phi)\mid \Box_i\phi\mid \bDia{i}{c}(\phi,\phi)$$
\noindent which allow formulas $\bDia{i}{c}(\phi, \psi)$ where $\phi\not=\psi.$ $\bDia{i}{c}(\phi, \psi)$ intuitively says that there are two $i$-successors such that one satisfies $\phi$ and the other satisfies $\psi$ and they do not agree on the value of $c$. The semantics is now standard for a binary modality: 
$$
\begin{array}{|lll|}
\hline
\M,s\vDash\bDia{i}{c}(\phi,\psi)& \iff& \exists u,v: \text{ such that }  sR_{i}^{c}uv, \M,u\vDash\phi \text{ and } \M,v\vDash\psi\\
\hline
\end{array}
$$
Surprisingly, the above language \MLKvrb\ is equally expressive as \MLKvr\ under the key observation by \cite{GW16} that $\bDia{i}{c}(\phi, \psi)$ is equivalent to the to the disjunction of the following three formulas: 
\begin{enumerate}
\item $\bDia{i}{c}\phi\land\Dia_i\psi$
\item $\bDia{i}{c}\psi\land\Dia_i\phi$
\item $\Dia_i\phi\land\Dia_i\psi\land\neg\bDia{i}{c}\phi\land\neg\bDia{i}{c}\psi\land\bDia{i}{c}(\phi\vee\psi)$
\end{enumerate}
Now it is clear that $\MLKvrb$ over Kripke models with binary and ternary relations is just a normal modal logic,  which also means that $\MLKvr$ (and thus $\ELKvr$) can be viewed as a disguised normal modal logic qua expressivity. Now the axiomatization and other technical issues can be largely simplified by using  standard techniques. \cite{GW16} showed the completeness of the following normal modal logic system using standard techniques,\footnote{Note that here the maximal consistent sets are enough to build the canonical model due to the change of models, compared to canonical model for \ELKvr.}  where $\EQUIV$, $\EXIST$ and $\DISBK$ capture exactly the three properties respectively.\footnote{Due to $\EQUIV$, we only need $\texttt{DISTKv}^b$ and $\texttt{NECKv}^b$ w.r.t. the first argument.}

\begin{minipage}{0.64\textwidth}
\begin{center}
\begin{tabular}{lc}
\multicolumn{2}{c}{System $\SMLKvb$}\\
\multicolumn{2}{l}{Axiom Schemas}\\
\texttt{TAUT} & \tr{ all the instances of tautologies}\\
\DISTK &$ \Box_i (p\to q)\to (\Box_i p\to \Box_i q)$\\
$\texttt{DISTKv}^b$ &$ \bBox{i}{c}(p\to q,r)\to (\bBox{i}{c}(p,r)\to \bBox{i}{c}(q,r))$\\
$\EQUIV$ &$\bBox{i}{c}(p,q)\to\bBox{i}{c}(q,p)$\\
$\EXIST$ &$\bDia{i}{c}(p,q)\to\Dia_i p$\\
$\DISBK$ &$\bDia{i}{c}(p,q)\land\Diamond_i r \to  \bDia{i}{c}(p,r)\lor \bDia{i}{c}(q,r)$\\
\end{tabular}
\end{center}
\end{minipage}
\hfill
\begin{minipage}{0.30\textwidth}
\begin{center}
\begin{tabular}{lc}
\multicolumn{2}{l}{Rules}\\
\texttt{MP}& $\dfrac{\phi,\phi\to\psi}{\psi}$\\
\GENK &$\dfrac{\phi}{\Box_i \phi}$\\
$\texttt{NECKv}^b$ &$\dfrac{\phi}{\bBox{i}{c}(\phi,\psi)}$\\
\SUB &$\dfrac{\phi}{\phi [p\slash\psi]}$\\
\end{tabular}
\end{center}
\end{minipage}\\

This normal modal logic view also gives us a standard bisimulation notion for $\MLKvrb$ on models with ternary and binary relations (cf. e.g., \cite{mlbook}). Then we can translate the bisimulation conditions on $R_{i}^{c}$ back to the conditions on $\to_i$ and the value assignment $V_\C$ to obtain a notion of bisimulation in the setting of FO epistemic models for \ELKvr. As another potential application, we believe that the normal modal logic view can also shed some light on the decision procedure of \ELKvr, since the models of $\MLKvr$ are free of value assignments, which are much easier to handle. 




\subsubsection{A new update operator}
We close the discussion on knowing value logic by another natural extension, which brings a surprising connection to dependence logic. So far, the updates we have considered are mainly public announcements. However, such updates are most suitable for changing knowledge-that. Actually, the knowing value operator $\Kv_i$ has also a very natural corresponding update operation. \cite{GvEW16} enrich \ELKvr\ with the \textit{public inspection} operator $[c]$ (call the following language \PILKvr): 
$$\phi::=\top\mid p\mid \neg\phi\mid (\phi\wedge\phi)\mid \K_i\phi\mid \Kv_i (\phi, c)\mid [c]\phi$$
Intuitively, $[c]\phi$ says that after revealing the \textit{actual} value of $c$, $\phi$ holds. It can be viewed as the knowing value analog of the public announcement of a formula. Formally, the semantics of $[c]\phi$ is defined on first-order epistemic models $\M=\langle S, D, \simall, V, V_{\D}\rangle$ as follows:
$$
\begin{array}{|lll|}
\hline
\M,s\vDash [c]\phi & \Leftrightarrow & \M|^{s}_c,s\vDash\phi\\
\hline
\end{array}
$$
where $\M|^s_c=\langle S', D, \{\sim_i|_{S'\times S'}\mid i\in \Ag\}, V|_{S'}, V_{\D}|_{\C\times S'}\rangle$ where $S'=\{s'\mid V_{\D}(c,s')=V_{\D}(c,s) \}$ i.e., the update deletes the worlds which do not agree with the current world $s$ on the value of $c$. In contrast to the update of public announcement, the update here is \textit{local} in the sense that the $s$ matters in the updated model $\M|^s_c$. 

By adapting some suitable bisimulation notion, \cite{GvEW16} showed that $\PILKvr$ is more expressive than $\ELKvr$, thus $[c]$ is not reducible. Intuitively, the update $[c]$ may bring new information that is not pre-encoded by a formula.

Now with this new dynamic operator at hand, we can express the knowledge of dependence between different constants as $\Kd_i(c, d):= \K_i[c]\Kv_id$. $\Kd_i(c, d)$ intuitively says that agent $i$ knows that the value of $d$ depends on the value of $c$. Formally the semantics can be spelled out:  
$$
\begin{array}{|lll|}
\hline
\M,s\vDash Kd_i(c, d) & \Leftrightarrow & \text{ for all }t_1\sim_is, t_2\sim_i s: t_1=_{c} t_2 \implies  t_1=_{d} t_2\\
\hline
\end{array}
$$
\noindent where $t=_c t'$ iff $V_\D(c, t)=V_\D(c, t')$. It is not hard to see that  $\Kd_i(c,d) \land  \Kv_i(\phi, c) \to \Kv_i(\phi, d)$ is valid: knowing the dependence helps to know the value. Moreover, we can handle the knowledge of dependence between sets of constants. Given any two finite sets $D, E \subseteq \C$ such that $D=\{d_1,\dots, d_n\}$ and $E=\{e_1,\dots e_m\}$, let $Kd_i(D,E):= K_i[d_1]\ldots[d_n](\Kv_i e_1 \land \ldots \land \Kv_i e_m)$. Note that the order of public inspections does not really matter.\footnote{A similar operator with propositional arguments was proposed by \cite{GK15} in the  setting of knowing whether, which can express that given the truth values of $\phi_1,\dots,\phi_n$ the agent $i$ knows whether $\phi$.}  

\medskip

$\Kd_i(c, d)$ can be viewed as the atomic formula $=\!\!(c, d)$ in \textit{dependence logic} proposed by \cite{Depbook}, w.r.t.\ the ``team'' model which consists of the $i$-accessible worlds (as value assignments for constants in $\C$). Note that there is a crucial difference between our approach and the team semantics of dependence atoms in dependence logic. We can specify the \textit{local} dependence by $[c]\Kv_id$ i.e., $i$ knows the value of $d$ given the actual value of $c$, whereas $=\!\!(c, d)$ can only specify \textit{global} dependence as the distinction between $\K_i[c]\Kv_id$ and $[c]\Kv_id$ shows. The connection with dependence logic also bring \PILKvr\ closer to the first-order variant of the \textit{epistemic inquisitive logic} by \cite{CiardelliR15}, where the knowledge of entailment of interrogatives can also be viewed as our $\Kd_i(c, d)$. More precisely, $\Kd_i(c, d)$ can be expressed by $K_i(\bar{\exists}x (x=c)\to \bar{\exists}x (x=d))$, where $\bar{\exists}$ is the inquisitive existential quantifier and $\bar{\exists}x (x=c)$ corresponds to the question on the value of $c$. Intuitively, $K_i(\bar{\exists}x (x=c)\to \bar{\exists}x (x=d))$ says that agent $i$ knows that the answer to the question ``what is $c$?'' will determine the answer to the question ``what is $d$?'', see \cite[Sec. 6.7.4]{Ivano16} for a detailed comparison with our approach.  

\cite{GvEW16} axiomatized the following single-agent fragment of \PILKvr\ which can be considered as the $\Kv$ counterpart of the public announcement logic: 
 \[ \phi ::= \top  \mid  \lnot\phi \mid \phi\land\phi \mid \Kv   ~c \mid [c]\phi \] 
However the axiomatization of the full \PILKvr\ is still open. On the other hand, \cite{Baltag16} proposed a very general language with a similar conditional operator $K_i^{\{t_1,\dots,t_n\}}$ where $t_i$ are terms that can contain function symbols over variables and formulas. $K_i^{\{t_1,\dots,t_n\}}t$ says that $i$ knows the value of $t$ if he or she is given the values of $t_1,\dots,t_n$. A distinct feature of this language, compared to the \Kv-based languages, is that it also includes equalities of terms as atomic formulas in order to obtain a complete axiomatization. It is shown that this language can pre-encode the public inspection operators and it is decidable.

\subsection{Knowing how}Last but not the least, we will look at a logic of a particular kind of knowing how proposed and studied by \cite{Wang15:lori}.\footnote{The extended full version will appear as \cite{Wang17}.} Compared to the previous two cases, it has a couple of special features worth mentioning: 
\begin{itemize}
\item There is no consensus on the logical language and the semantics of the logic of knowing how. 
\item As we will see, although the knowing how formulas still follow roughly the general shape of $\exists x \Box \phi(x)$, the existential quantifier is not really a first-order one. 
\item Contrary to the previous cases of knowing whether and knowing what, there can be \textit{more} than one $x$ that can make $\Box \phi(x)$ true in the knowing how case, and this requires new techniques in the completeness proof. 
\item Our model is no longer based on epistemic models with epistemic relations. 
\end{itemize}

\medskip

Knowing how is frequently discussed in epistemology and in AI.\footnote{See \citep{Wang15:lori} for a more detailed survey.} Philosophers debate about whether knowledge-how, the knowledge expressed by the knowing how expressions, can be reduced to knowledge-that, i.e. propositional knowledge.\footnote{See the collection of papers on the topic at philpaper edited by by John Bengson: \url{http://philpapers.org/browse/knowledge-how}} There are two major philosophical stances: \textit{intellectualists} think knowledge-how is reducible to knowledge-that (cf. e.g., \cite{SW10}), while \textit{anti-intellectualists} holds the opposite position that knowledge-how is irreducible (cf e.g., \cite{Ryle49}). At the first glance, knowing how seems to express a statement about ability, e.g., ``I know how to swim'' roughly says that I have the ability to swim. However, philosophy literature provides ample examples to show that this simple-minded idea is shaky, e.g., can you say you know how to digest food since you have that ability? As another example, in some cases even though you do not have the ability at the moment, it is still reasonable to claim the knowledge-how, e.g., a pianist with a broken-arm may still say he or she knows how to play piano, although due to the accident he or she cannot do it right now.\footnote{Such examples motivated intellectualists to propose an account other than treating knowledge-how simply as ability. A notable approach proposed by \cite{SW10} breaks down ``knowing how to $F$'' into: ``There is a way such that I know it is a way to do F, and I entertain it in a \textit{practical mode of presentation}.'' Note that it essentially has the familiar shape $\exists x\K \phi(x)$,  which also inspired the semi-formal treatment by \cite{LauWang}.} Here the relevant insight is that knowing how expressions may come with implicit conditions. When we say that a chef knows how to cook Chinese dishes, it does not mean that he can do it right now, but it means he can do it \textit{given} all the ingredients and facilities. Thus in the formal language we introduce a binary modality $\Kh(\psi, \phi)$ meaning that I know how to achieve $\phi$ given $\psi.$ Note that $\psi$ may be false currently but we should look at all the worlds where it is true. 

In AI, ever since the pioneering works by \cite{McCarthy79} and \cite{Moore77}, formalizing the interaction of knowledge and ability has been an important issue till now (cf. \citep{Gochet13,KandA15} for up-to-date overviews). 
One problem that logicians in AI face is that simply combining ``knowing that'' and ``ability'' does not lead to a natural notion of knowing how, as sharply pointed out by \cite{Herzig15}. For example, adding the knowing-that operator to \textit{alternating temporal logic} (\textbf{ATL}) can result in a logic which can express one knows that there is a strategy to achieve some goal, which is in the \textit{de dicto} shape of $\K\exists x \phi(x)$ rather than the desired \textit{de re} shape $\exists x K \phi(x)$. We need a way to somehow insert the $\K$ modality in-between the implicit existential quantifier and the strategy modality.\footnote{See \citep{Herzig15} for some  existing solutions, e.g. by using epistemic STIT logic proposed by  \cite{Broersen2015}.} We tackle this problem by packing the quantifier and the modality together in the  $\Kh$ operator with a semantics inspired by \textit{conformant planning} in AI, where the goal is to find a uniform plan (action sequence) such that at all the initial situations the plan will always work and reach the goal (cf. \cite{YLW15}). Knowing how to achieve $\phi$ given $\psi$ then amounts to having a conformant plan which works for all the $\psi$-worlds. 

Before going into the details,  some clarifications have to be made. 
\begin{itemize}
\item We only focus on \textit{goal-directed} knowing-how, as \cite{Gochet13} puts it, e.g., knowing how to prove a theorem,  how to open the door, how to bake a cake, and how to cure the disease.  
\item We do not study knowing-how in the following senses: I know how the computer works (explanation); I know how happy she is (degree of emotion); I know how to behave at the dinner table (rule-directed).     
\end{itemize}

\subsubsection{Language and semantics}
As inspired by the philosophy literature,
we introduce a conditional knowing-how operator in the following single-agent language $\LKh$ \citep{Wang15:lori}:
$$\begin{array}{r@{\quad::= \quad}l}
\varphi  &
  \top
         \mid
           p
           \mid \neg \varphi
           \mid (\varphi \land \varphi)
           \mid \Kh (\varphi, \varphi)
\end{array}$$
Intuitively, $\Kh (\psi, \varphi)$ says that the agent knows how to achieve $\phi$ given the condition $\psi$.
$\U\varphi$ is defined as $ \Kh(\neg\varphi,\bot)$, which is intended to be a \textit{universal modality} to be explained later.  

Given a non-empty set of propositional letters $\BP$, a non-empty set of actions $\Act$, a model is simply a tuple $(S, R, V)$ where: 
\begin{itemize}
\item $S$ is a non-empty set of states;
\item $R: \Act\to 2^{S\times S}$ is a collection of transitions labelled by actions in $\Act$;
\item $V: S\to 2^{\BP}$ is a valuation function.
\end{itemize}
Note that this is \textit{not} a standard epistemic model for there is no epistemic alternative relation in the model. Intuitively, the model represents the ability that the agent has, and it can be used as a model for an epistemic logic of knowing how (cf. also \citep{Wang15:icla} for a more general setting.). For example, the left model below represents that the agent can do $a$ on $s_1$ but he cannot control the outcome. On the other hand he can do $b$ on $s_2$ which leads to a single $q$-world. 
 
$$
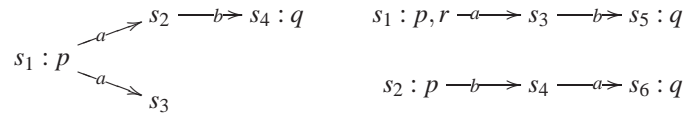
\xymatrix@R-20pt{
&{s_2}\ar[r]|b&{s_4: q}\\
{s_1:p}\ar[ur]|a\ar[dr]|a\\
&{s_3}
\restore
}
\qquad 
\xymatrix@R-10pt{
{s_1:p,r}\ar[r]|a&{s_3}\ar[r]|b&{s_5: q}\\
{s_2:p}\ar[r]|b&{s_4}\ar[r]|a&{s_6: q}
\restore
}
$$
Intuitively, given only $p$, the agent should not know how to reach $q$ in the above two models: although $ab$ leads to $q$ in the left model, $a$ cannot control the result of $a$; he may fail to continue to do $b$ after doing $a$. For the right model, although the agent can do $ab$ to reach $q$ on $s_1$ and do $ba$ to reach $q$ on $s_2$, he does not know where he is exactly given only $p$, and thus does not have a \textit{uniform} plan which can always work. We flesh out such intuition in the following semantics:  
$$\small{\begin{array}{|rcl|}
\hline
\M,s\vDash \Kh(\psi,\varphi)&\Leftrightarrow& \text{ \textit{there exists} an action sequence } \sigma\in\Act^* \text{ such that \textit{for all} } \M, s'\vDash \psi: \\
&& (1) \ \sigma \text{ is \textit{strongly executable} at $s'$, and } \\
&& (2) \text{\textit{ for all} $t$ if } s'\rel{\sigma}t \text{ then }\M, t\vDash \varphi    \\
\hline
\end{array}}
$$
\noindent where $\sigma=a_1\dots a_n$ is \emph{strongly executable} at $s'$ if $s'$ has at least one $a_1$-successor and for any $1\leq k < n$ and any $t$, $s'\rel{a_1\dots a_k}t$ implies that $t$ has at least one $a_{k+1}$-successor. Intuitively, $\sigma$ is strongly executable iff  $\sigma$ is executable and whenever you start doing an initial segment of $\sigma$, you can always continue. For example $ab$ is not strongly executable at $s_1$ in the left model above, since it may fail. Note that the quantifier schema in the semantics is $\exists \forall$ which is in compliance with the general schema $\exists x \K$, although now the existential quantifier is no longer first-order, and the $\K$ is replaced by a quantifier induced by the condition $\psi$ representing the initial uncertainty. 

One can verify that $s_1\vDash\neg\Kh(p, q)$ in the above two models and $s_1\vDash \Kh(p, q)$ in the model below, since there is a strongly executable plan $ru$ from any $p$-world to some $q$-world. 
$$
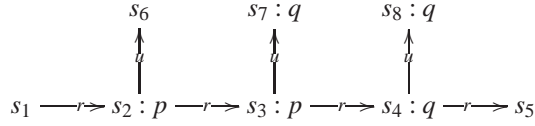
\xymatrix{
&s_6&{{s_7:q}}&{{s_8: q}} &\\
s_1\ar[r]|r& s_2:p\ar[r]|r\ar[u]|u& s_3:p\ar[r]|r\ar[u]|u&{s_4:q}\ar[r]|r\ar[u]|u&s_5
\restore
}$$

Now it can also be verified that $U$ is indeed a universal modality: 
$$\begin{array}{|rcrcl|}
\hline

\M,s\vDash \U \varphi&\Leftrightarrow &\Kh(\neg\phi, \bot)&\Leftrightarrow& \text{ for all }t\in \S, \M, t\vDash\varphi   \\
\hline
\end{array}
$$

\subsubsection{Axiomatization}
A complete axiomatization is given in \citep{Wang15:lori} using $\Kh$ and the definable $\U$: 
\begin{center}
\begin{tabular}{lclc}
\multicolumn{4}{c}{System $\SKh$}\\
\multicolumn{2}{l}{\textbf{Axioms}}&\textbf{Rules}&\\
\TAUT & \text{all axioms of propositional logic}&\MP & $\dfrac{\varphi,\varphi\to\psi}{\psi}$\\
\DISTU & $\U p\land\U (p\to q)\to \U q$&\NECU &$\dfrac{\varphi}{\U\varphi}$\\
\COMPKh & $\Kh(p, r)\land\Kh(r, q)\to\Kh(p, q)$&\SUB &$\dfrac{\varphi(p)}{\varphi[\psi\slash p]}$\\
\EMP &$\U(p \to q)\to \Kh(p, q)$  &\phantom{$\dfrac{\varphi}{[a]\varphi} $}\\
\AxTrU& $\U p\to p $ &\phantom{$\dfrac{\varphi(p)}{\varphi[\psi\slash p]}$} \\
 \AxTransKU& $\Kh(p, q)\to\U\Kh(p, q)$&\phantom{$\dfrac{\varphi}{\Box\varphi}$}\\
 \AxEucKU& $\neg \Kh(p, q)\to\U\neg\Kh(p, q)$&\phantom{$\dfrac{\varphi}{\Box\varphi}$}\\
\end{tabular}
\end{center}
We can view $\U$ as a knowing that operator for the background knowledge taken for granted in the model, and it indeed behaves as an $\mathbb{S}5$ modality.\footnote{We can derive $\U p\to \U\U p$ and $\neg \U p\to U\neg \U p$ \citep{Wang15:lori}.} \AxEucKU\ and  \AxTransKU\ are the introspection axioms. \EMP\ says that if you know $p$ implies $q$ then you trivially know how to achieve $q$ given $p$, i.e., doing nothing. The most interesting axiom is $\COMPKh$, which says knowledge-how can be sequentially composed. Moreover, two interesting axioms below can be derived from the above system. \WSKh\ says that you can strengthen the precondition and weaken the goal and still know how; \POSTKh\ is a recursive way of expressing the compositionality of knowing-how. 
\begin{center}
\begin{tabular}{|c|c|}
\hline
\WSKh & $ \U(p\to r)\land \U(o\to q)\land \Kh(r, o)\to \Kh(p, q)$\\
\POSTKh & $\Kh(r, \Kh(p, q)\land p)\to \Kh(r, q) $\\
 \hline
\end{tabular}
\end{center}
\begin{theorem}[\cite{Wang15:lori}]
$\SKh$ is sound and strongly complete w.r.t. the class of all models. 
\end{theorem}
The completeness proof involves building special canonical models,\footnote{For each maximal consistent set we build a canonical model \citep{Wang15:lori}.} where every $\Kh(\psi, \phi)$ can be realized by a simple one-step simple plan. Note that in contrast with the previous logics of knowing whether and knowing value, when showing $\neg\Kh(\psi, \phi)$ is true at a maximal consistent set including it, it is no longer enough to build two differentiating states, since the existential quantifier hidden in $\Kh$ no longer assumes uniqueness: there can be many plans to achieve $\phi$ on a given $\psi$-world.\footnote{Recall that $\neg \Kv_i c$ is true if there are two states which disagree on $c$.} However, you need to show no \textit{single} plan will do the job uniformly over all the $\psi$-worlds.   

In a canonical model, all the states share the same $\Kh$-formulas, it is then easy to prove that the size of the canonical model is bounded by $2^n$ where $n$ is the number of propositional letters. Therefore for a given $\LKh$ formula $\phi$, if it is satisfiable then it is satisfiable in a model which is bounded by $2^{|\phi|}$. This leads to the small model property of the logic, and the decidability follows since we have a finite axiomatization, as shown by \cite{Wang17}. 

It is also natural to generalize the $\Kh$ operator to a ternary one with an extra intermediate constraint. $\Kh(\psi, \chi,\phi)$ then says that ``the agent knows how to achieve $\phi$ given $\psi$ while maintaining $\chi$ in-between.'' In this way we can handle knowledge-how with constants about the process of the plan. The logic of this ternary modality is formally characterized by \cite{LiWang17}. 

\medskip

Having presented our examples of the logics of knowing whether, knowing what, and knowing how, we encourage the readers to go back to the summary of the highlights about each logic at the beginning of Section~\ref{sec.example}.

\section{Conclusions and future work}
This paper advocates the study of epistemic logics of knowing-wh. We started with a survey on Hintikka's contributions to knowing-wh, and the relevant recent literature on quantified epistemic logic. Then we proposed a new approach to epistemic logics of knowing-wh, which takes each knowing-wh as a single modality. In this way we can ``hide'' the quantifiers inside modalities, thus limiting the expressivity of the language in order to avoid conceptual and technical problems of the full quantified epistemic logic. By three example studies on knowing whether, knowing what and knowing how, we demonstrated the usefulness and the diversity of knowing-wh logics. We hope we have shown that this new approach may lead us to: 
\begin{itemize}
\item interesting (non-normal) modal operators packaging a quantifier and a (standard) modality ($\exists x \Box$);
\item new meaningful axioms about different knowing-wh and their interactions with the knowing that operator; 
\item discovery of computationally (relatively) cheap fragments of first-order or higher-order modal logics;
\item interesting connections with existing logics;
\item various techniques handling the completeness proof of such non-normal modal logics;
\item techniques restoring the symmetry between a weak language and rich models.
\end{itemize}

In some sense, our approach is a minimalistic one. We do not have the ambition to fit everything about knowing-wh in a very powerful language with full compositionality and the flexibility to capture the context-sensitivity. Instead, we start from very simple languages of some particular knowing-wh constructions, fix some intuitive semantics which can account for some useful readings, and then see whether we can capture the decidable logics nicely. Essentially, we are following the successful story of propositional modal logic, which packages quantifiers and other constructions together in modalities. This minimalistic idea distinguishes us from the quantified epistemic logic approach by Hintikka and others, and the linguistically motivated inquisitive semantics approach to the logic of knowing-wh. Our examples also showed that although the hidden  logical structures of various knowing-wh modalities may be similar to each other to some extent, the details of the language, models, and the semantics matter a lot in deciding the concrete axioms for different knowing-wh. The newly introduced modalities also let us see clearly the special features of different knowing-wh, which may not be possible if we break everything down into quantifiers, predicates, and standard modalities in a quantified epistemic logic.    

Having said the above, we are also aware of the obvious limitations of our approach. Readers are encouraged to go back to Section~\ref{sec.ELwh} to review the discussion on the advantages and limitations of our approach. We think both the minimalistic approach and the ``maximalistic'' approaches are good for their own purposes, and the two approaches can be beneficial for each other by bringing new insights to balance expressive power and complexity further. 

\medskip

We believe this is only the beginning of an exciting story. Besides the epistemic logics of  other types of knowing-wh such as knowing why \citep{Xu16} and knowing who and so on, there are a lot of general topics to be discussed about the existing logics mentioned in this paper. For example:
\begin{itemize}
\item model theory, proof theory, and complexity of the knowing-wh logics;
\item group notions of knowing-wh, e.g., commonly knowing whether, jointly knowing how and so on; 
\item new update mechanisms to change knowing-wh, e.g., learning new abilities in the model of knowing how;
\item simplified semantics, e.g., new semantics of knowing how logic that can keep the valid axioms intact but restores the symmetry between syntax and semantics, as in the case of knowing value logic. 
\item alternative semantics, e.g., multi-agent, contingent planning based knowing how logic, where branching plans are used; 
\item logical omniscience of knowing-wh;
\item the study of the generic modality which packs $\exists x \Box$ together, and its connection to monodic  and other decidable fragments of quantified modal logic.
\end{itemize}
This new generation of epistemic logics will open up various opportunities for epistemic logicians to explore.\footnote{In Hintikka's terms,  maybe it can be called the 1.5th generation of epistemic logics, since it is not as general as Hintikka's idea of the second generation epistemic logics.}

\begin{acknowledgement}
The author acknowledges the support from the National Program for Special Support of Eminent Professionals and NSSF key projects 12\&ZD119. The author is grateful to Hans van Ditmarsch for his very detailed comments on an early version of this paper. The author also thanks the anonymous reviewer who gave many constructive suggestions including the observation in footnote \ref{foot}. 
\end{acknowledgement}

%






\bibliography{BKT}

\begin{thebibliography}{108}
\providecommand{\natexlab}[1]{#1}
\providecommand{\url}[1]{{#1}}
\providecommand{\urlprefix}{URL }
\expandafter\ifx\csname urlstyle\endcsname\relax
  \providecommand{\doi}[1]{DOI~\discretionary{}{}{}#1}\else
  \providecommand{\doi}{DOI~\discretionary{}{}{}\begingroup
  \urlstyle{rm}\Url}\fi
\providecommand{\eprint}[2][]{\url{#2}}

\bibitem[{{\AA}gotnes et~al(2015){\AA}gotnes, Goranko, Jamroga, and
  Wooldridge}]{KandA15}
{\AA}gotnes T, Goranko V, Jamroga W, Wooldridge M (2015) Knowledge and ability.
  In: van Ditmarsch H, Halpern J, van~der Hoek W, Kooi B (eds) Handbook of
  Epistemic Logic, College Publications, chap~11, pp 543--589

\bibitem[{Aloni(2001)}]{Aloni01}
Aloni M (2001) Quantification under conceptual covers. PhD thesis, University
  of Amsterdam

\bibitem[{Aloni(2016)}]{Aloni16}
Aloni M (2016) Knowing-who in quantified epistemic logic. In: van Ditmarsch H,
  Sandu G (eds) Jaakko Hintikka on knowledge and game theoretical semantics,
  Springer

\bibitem[{Aloni and Roelofsen(2011)}]{aloniinterpreting}
Aloni M, Roelofsen F (2011) Interpreting concealed questions. Linguistics and
  Philosophy 34(5):443--478

\bibitem[{Aloni et~al(2013)Aloni, {\'{E}}gr{\'{e}}, and de~Jager}]{AloniEJ13}
Aloni M, {\'{E}}gr{\'{e}} P, de~Jager T (2013) Knowing whether {A} or {B}.
  Synthese 190(14):2595--2621

\bibitem[{Aumann(1989)}]{Aumann89}
Aumann R (1989) Notes on interactive epistemology. In: Cowles Foundation for
  Reaserch in Economics working paper

\bibitem[{Baltag(2016)}]{Baltag16}
Baltag A (2016) To know is to know the value of a variable. In: Proceedings of
  AiML Vol. 11, pp 135--155

\bibitem[{Belardinelli and van~der Hoek(2015)}]{Belardinelli:2015}
Belardinelli F, van~der Hoek W (2015) Epistemic quantified boolean logic:
  Expressiveness and completeness results. In: Proceedings of IJCAI '15, AAAI
  Press, pp 2748--2754

\bibitem[{Belardinelli and van~der Hoek(2016)}]{BelardinelliH16}
Belardinelli F, van~der Hoek W (2016) A semantical analysis of second-order
  propositional modal logic. In: Proceedings of AAAI'16, pp 886--892

\bibitem[{Belardinelli and Lomuscio(2009)}]{BelardinelliL09}
Belardinelli F, Lomuscio A (2009) Quantified epistemic logics for reasoning
  about knowledge in multi-agent systems. Artificial Intelligence
  173(9-10):982--1013

\bibitem[{Belardinelli and Lomuscio(2011)}]{BelardinelliL11}
Belardinelli F, Lomuscio A (2011) First-order linear-time epistemic logic with
  group knowledge: An axiomatisation of the monodic fragment. Fundamenta
  Informaticae 106(2-4):175--190

\bibitem[{Belardinelli and Lomuscio(2012)}]{BelardinelliL12}
Belardinelli F, Lomuscio A (2012) Interactions between knowledge and time in a
  first-order logic for multi-agent systems: Completeness results. Journal of
  Artificial Intelligence Research 45:1--45

\bibitem[{Blackburn et~al(2002)Blackburn, de~Rijke, and Venema}]{mlbook}
Blackburn P, de~Rijke M, Venema Y (2002) Modal Logic. {Cambridge University
  Press}

\bibitem[{Boer and Lycan(2003)}]{BL03}
Boer SE, Lycan WG (2003) Knowing Who. The MIT Press

\bibitem[{Boja{\'{n}}czyk(2013)}]{Bojańczyk2013}
Boja{\'{n}}czyk M (2013) Modelling infinite structures with atoms. In: Libkin
  L, Kohlenbach U, de~Queiroz R (eds) Proceedings of WoLLIC'13, Springer, pp
  13--28

\bibitem[{Boja\'{n}czyk et~al(2011)Boja\'{n}czyk, David, Muscholl, Schwentick,
  and Segoufin}]{Bojanczyk:2011}
Boja\'{n}czyk M, David C, Muscholl A, Schwentick T, Segoufin L (2011)
  Two-variable logic on data words. ACM Transactions on Computational Logic
  12(4):27:1--27:26

\bibitem[{Bra\"uner and Ghilardi(2007)}]{G07FML}
Bra\"uner T, Ghilardi S (2007) First-order modal logic. In: Blackburn P, van
  Benthem J, Wolter F (eds) Handbook of Modal Logic, pp 549--620

\bibitem[{Broersen and Herzig(2015)}]{Broersen2015}
Broersen J, Herzig A (2015) Using \text{STIT} theory to talk about strategies.
  In: van Benthem J, Ghosh S, Verbrugge R (eds) Models of Strategic Reasoning:
  Logics, Games, and Communities, Springer, pp 137--173

\bibitem[{Chaum(1988)}]{DC}
Chaum D (1988) The dining cryptographers problem: Unconditional sender and
  recipient untraceability. Journal of Cryptology 1(1):65--75

\bibitem[{Ciardelli(2014)}]{Ciardelli14}
Ciardelli I (2014) Modalities in the realm of questions: Axiomatizing
  inquisitive epistemic logic. In: Proceedings of AiML Vol. 10, pp 94--113

\bibitem[{Ciardelli(2016)}]{Ivano16}
Ciardelli I (2016) Questions in logic. PhD thesis, University of Amsterdam

\bibitem[{Ciardelli and Roelofsen(2015)}]{CiardelliR15}
Ciardelli I, Roelofsen F (2015) Inquisitive dynamic epistemic logic. Synthese
  192(6):1643--1687

\bibitem[{Ciardelli et~al(2013)Ciardelli, Groenendijk, and
  Roelofsen}]{CiardelliGR13}
Ciardelli I, Groenendijk J, Roelofsen F (2013) Inquisitive semantics: {A} new
  notion of meaning. Language and Linguistics Compass 7(9):459--476

\bibitem[{Cohen and Dam(2007)}]{CoDa07}
Cohen M, Dam M (2007) A complete axiomatization of knowledge and cryptography.
  In: Proceedings of LiCS '07, IEEE Computer Society, pp 77--88

\bibitem[{Corsi(2002)}]{Corsi02}
Corsi G (2002) A unified completeness theorem for quantified modal logics.
  Journal of Symbolic Logic 67(4):1483--1510

\bibitem[{Corsi and Orlandelli(2013)}]{CorsiO13}
Corsi G, Orlandelli E (2013) Free quantified epistemic logics. Studia Logica
  101(6):1159--1183

\bibitem[{Corsi and Tassi(2014)}]{CorsiT14}
Corsi G, Tassi G (2014) A new approach to epistemic logic. In: Logic,
  Reasoning, and Rationality, Springer, pp 25--41

\bibitem[{Cresswell(1988)}]{cresswell1988necessity}
Cresswell MJ (1988) Necessity and contingency. Studia Logica 47(2):145--149

\bibitem[{Demri(1997)}]{Demri97}
Demri S (1997) A completeness proof for a logic with an alternative necessity
  operator. Studia Logica 58(1):99--112

\bibitem[{Ding(2015)}]{Ding15}
Ding Y (2015) Axiomatization and complexity of modal logic with knowing-what
  operator on model class \text{K},
  \urlprefix\url{http://www.voidprove.com/research.html}, unpublished
  manuscript

\bibitem[{van Ditmarsch(2007)}]{Dit07:comments}
van Ditmarsch H (2007) Comments to 'logics of public communications'. Synthese
  158(2):181--187

\bibitem[{van Ditmarsch and Fan(2016)}]{vDF16}
van Ditmarsch H, Fan J (2016) Propositional quantication in logics of
  contingency. Journal of Applied Non-Classical Logics (forthcoming)

\bibitem[{van Ditmarsch et~al(2007)van Ditmarsch, van~der Hoek, and
  Kooi}]{DELbook}
van Ditmarsch H, van~der Hoek W, Kooi B (2007) Dynamic Epistemic Logic.
  Springer

\bibitem[{van Ditmarsch et~al(2012{\natexlab{a}})van Ditmarsch, van~der Hoek,
  and Illiev}]{vDHI}
van Ditmarsch H, van~der Hoek W, Illiev P (2012{\natexlab{a}}) Everything is
  knowable – how to get to know whether a proposition is true. Theoria 78(2)

\bibitem[{van Ditmarsch et~al(2012{\natexlab{b}})van Ditmarsch, van~der Hoek,
  and Kooi}]{vanDitmarsch2012}
van Ditmarsch H, van~der Hoek W, Kooi B (2012{\natexlab{b}}) Local properties
  in modal logic. Artificial Intelligence 187–-188:133 -- 155

\bibitem[{van Ditmarsch et~al(2014)van Ditmarsch, Fan, van~der Hoek, and
  Iliev}]{DFHI14}
van Ditmarsch H, Fan J, van~der Hoek W, Iliev P (2014) Some exponential lower
  bounds on formula-size in modal logic. In: Proceedings of AiML Vol.10, pp
  139--157

\bibitem[{van Ditmarsch et~al(2015)van Ditmarsch, Halpern, van~der Hoek, and
  Kooi}]{ELbook}
van Ditmarsch H, Halpern J, van~der Hoek W, Kooi B (eds)  (2015) Handbook of
  Epistemic Logic. College Publications

\bibitem[{Egr\'e(2008)}]{Egre08}
Egr\'e P (2008) Question-embedding and factivity. Grazer Philosophische Studien
  77(1):85--125

\bibitem[{Fagin et~al(1995)Fagin, Halpern, Moses, and Vardi}]{RAK}
Fagin R, Halpern J, Moses Y, Vardi M (1995) Reasoning about knowledge. MIT
  Press, Cambridge, MA, USA

\bibitem[{Fan(2015)}]{FanPhD}
Fan J (2015) Logical studies for non-contingency operator. PhD thesis, Peking
  University, (in Chinese)

\bibitem[{Fan and van Ditmarsch(2015)}]{FanD15}
Fan J, van Ditmarsch H (2015) Neighborhood contingency logic. In: Proceedings
  of ICLA'15, pp 88--99

\bibitem[{Fan et~al(2014)Fan, Wang, and van Ditmarsch}]{FWvD14}
Fan J, Wang Y, van Ditmarsch H (2014) Almost neccessary. In: Proceedings of
  AiML Vol.10, pp 178--196

\bibitem[{Fan et~al(2015)Fan, Wang, and van Ditmarsch}]{FWvD15}
Fan J, Wang Y, van Ditmarsch H (2015) Contingency and knowing whether. The
  Review of Symbolic Logic 8:75--107

\bibitem[{{Fine}(1970)}]{Fine1970}
{Fine} K (1970) Propositional quantifiers in modal logic. Theoria
  36(3):336--346

\bibitem[{Fitch(1963)}]{fitch1963a}
Fitch F (1963) A logical analysis of some value concepts. Journal of Symbolic
  Logic 28(2):135--142

\bibitem[{Fitting and Mendelsohn(1998)}]{FittingM1998}
Fitting M, Mendelsohn RL (1998) First-Order Modal Logic. Springer

\bibitem[{van Fraassen(1980)}]{vanfraassen1980}
van Fraassen B (1980) The scientific image. Oxford: Oxford University Press

\bibitem[{Garson(2001)}]{Garson2001}
Garson JW (2001) Quantification in modal logic. In: Gabbay DM, Guenthner F
  (eds) Handbook of Philosophical Logic, Springer, Dordrecht, pp 267--323

\bibitem[{Gattinger et~al(2017)Gattinger, van Eijck, and Wang}]{GvEW16}
Gattinger M, van Eijck J, Wang Y (2017) Knowing values and public inspection.
  In: Proceedings of ICLA'17, forthcoming

\bibitem[{Gochet(2013)}]{Gochet13}
Gochet P (2013) An open problem in the logic of knowing how. In: Hintikka J
  (ed) Open Problems in Epistemology, The Philosophical Society of Finland

\bibitem[{Gochet and Gribomont(2006)}]{gochet2006epistemic}
Gochet P, Gribomont P (2006) Epistemic logic. In: Gabbay DM, Woods J (eds)
  Handbook of the History of Logic, vol~7

\bibitem[{Goranko and Kuusisto(2015)}]{GK15}
Goranko V, Kuusisto A (2015) Logics for propositional determinacy and
  independence, \urlprefix\url{https://arxiv.org/abs/1609.07398}, manuscript

\bibitem[{Gu and Wang(2016)}]{GW16}
Gu T, Wang Y (2016) ``knowing value'' logic as a normal modal logic. In:
  Proceedings of AiML Vol.11, 362--381

\bibitem[{Harrah(2002)}]{Harrah02}
Harrah D (2002) The logic of questions. In: Gabbay D (ed) Handbook of
  Philosophical Logic, vol~8

\bibitem[{Hart et~al(1996)Hart, Heifetz, and Samet}]{HHS96:knowingWT}
Hart S, Heifetz A, Samet D (1996) ``knowing whether'', ``knowing that'', and
  the cardinality of state spaces. Journal of Economic Theory 70(1):249--256

\bibitem[{Heim(1979)}]{heim1979}
Heim I (1979) Concealed questions. In: B\"{a}uerle R, Egli U, von Stechow A
  (eds) Semantics from Different Points of View, pp 51--60

\bibitem[{Herzig(2015)}]{Herzig15}
Herzig A (2015) Logics of knowledge and action: critical analysis and
  challenges. Autonomous Agents and Multi-Agent Systems 29(5):719--753

\bibitem[{Herzig et~al(2015)Herzig, Lorini, and Maffre}]{HerzigLM15}
Herzig A, Lorini E, Maffre F (2015) A poor man's epistemic logic based on
  propositional assignment and higher-order observation. In: Proceedings of
  LORI-V, pp 156--168

\bibitem[{Hintikka(1962)}]{Hintikka:kab}
Hintikka J (1962) Knowledge and Belief: An Introduction to the Logic of the Two
  Notions. Cornell University Press, Ithaca N.Y.

\bibitem[{Hintikka(1989{\natexlab{a}})}]{Hintikka1989b}
Hintikka J (1989{\natexlab{a}}) On sense, reference, and the objects of
  knowledge. In: The Logic of Epistemology and the Epistemology of Logic:
  Selected Essays, Springer, pp 45--61

\bibitem[{Hintikka(1989{\natexlab{b}})}]{Hintikka1989}
Hintikka J (1989{\natexlab{b}}) Reasoning about knowledge in philosophy: The
  paradigm of epistemic logic. In: The Logic of Epistemology and the
  Epistemology of Logic: Selected Essays, Springer, pp 17--35

\bibitem[{Hintikka(1996)}]{Hintikka95:KAK}
Hintikka J (1996) Knowledge acknowledged: Knowledge of propositions vs.
  knowledge of objects. Philosophy and Phenomenological Research 56(2):251--275

\bibitem[{Hintikka(1999)}]{Hintikka1999}
Hintikka J (1999) What is the logic of experimental inquiry? In: Inquiry as
  Inquiry: A Logic of Scientific Discovery, Springer, pp 143--160

\bibitem[{Hintikka(2003)}]{Hintikka2003}
Hintikka J (2003) A second generation epistemic logic and its general
  significance. In: Hendricks VF, J{\o}rgensen KF, Pedersen SA (eds) Knowledge
  Contributors, Springer, pp 33--55

\bibitem[{Hintikka(2007)}]{hintikka2007socratic}
Hintikka J (2007) Socratic epistemology: {Explorations} of knowledge-seeking by
  questioning. Cambridge: Cambridge University Press

\bibitem[{Hintikka and Halonen(1995)}]{Hintikka95}
Hintikka J, Halonen I (1995) Semantics and pragmatics for why-questions. The
  Journal of Philosophy 92(12):636--657

\bibitem[{Hintikka and Sandu(1989)}]{hintikkaSandu1989}
Hintikka J, Sandu G (1989) Informational independence as a semantical
  phenomenon. In: Fenstad JE, Frolov IT, Hilpinen R (eds) Logic, Methodology
  and Philosophy of Science 8, Elsevier, pp 571–--589

\bibitem[{Hintikka and Symons(2003)}]{HinSym2003}
Hintikka J, Symons J (2003) Systems of visual identification in neuroscience:
  Lessons from epistemic logic. Philosophy of Science 70(1):89--104

\bibitem[{Hodkinson(2002)}]{Hodkinson02a}
Hodkinson IM (2002) Monodic packed fragment with equality is decidable. Studia
  Logica 72(2):185--197

\bibitem[{Hodkinson et~al(2000)Hodkinson, Wolter, and
  Zakharyaschev}]{HodkinsonWZ00}
Hodkinson IM, Wolter F, Zakharyaschev M (2000) Decidable fragment of
  first-order temporal logics. Annals of Pure and Applied Logic
  106(1-3):85--134

\bibitem[{Hodkinson et~al(2002)Hodkinson, Wolter, and
  Zakharyaschev}]{HodkinsonWZ02}
Hodkinson IM, Wolter F, Zakharyaschev M (2002) Decidable and undecidable
  fragments of first-order branching temporal logics. In: Proceedings of
  LiCS'02, pp 393--402

\bibitem[{van~der Hoek and Lomuscio(2004)}]{hoeketal:2004}
van~der Hoek W, Lomuscio A (2004) A logic for ignorance. Electronic Notes in
  Theoretical Computer Science 85(2):117--133

\bibitem[{Holliday and Perry(2014)}]{HollidayP2014}
Holliday WH, Perry J (2014) Roles, rigidity, and quantification in epistemic
  logic. In: Baltag A, Smets S (eds) Johan van Benthem on Logic and Information
  Dynamics, Springer, pp 591--629

\bibitem[{Humberstone(1995)}]{Humberstone95}
Humberstone L (1995) The logic of non-contingency. Notre Dame Journal of Formal
  Logic 36(2):214--229

\bibitem[{Kaneko and Nagashima(1996)}]{KanekoN96}
Kaneko M, Nagashima T (1996) Game logic and its applications. Studia Logica
  57(2/3):325--354

\bibitem[{Khan and Banerjee(2010)}]{Khan2010}
Khan MA, Banerjee M (2010) A logic for multiple-source approximation systems
  with distributed knowledge base. Journal of Philosophical Logic
  40(5):663--692

\bibitem[{Kuhn(1995)}]{Kuhn95}
Kuhn S (1995) Minimal non-contingency logic. Notre Dame Journal of Formal Logic
  36(2):230--234

\bibitem[{Lau and Wang(2016)}]{LauWang}
Lau T, Wang Y (2016) Knowing your ability. The Philosophical Forum pp 415--424

\bibitem[{Lenzen(1978)}]{Lenzen78}
Lenzen W (1978) Recent work in epistemic logic. Acta Philosophica Fennica
  30(2):1--219

\bibitem[{Li and Wang(2017)}]{LiWang17}
Li Y, Wang Y (2017) Achieving while maintaining: A logic of knowing how with
  intermediate constraints. In: Proceedings of ICLA'17, forthcoming

\bibitem[{Liu and Wang(2013)}]{LiuW13}
Liu F, Wang Y (2013) Reasoning about agent types and the hardest logic puzzle
  ever. Minds and Machines 23(1):123--161

\bibitem[{Lomuscio and Ryan(1999)}]{LomuscioR99}
Lomuscio A, Ryan M (1999) A spectrum of modes of knowledge sharing between
  agents. In: Proceedings of ATAL'99, pp 13--26

\bibitem[{Ma and Guo(1983)}]{XiwenW83}
Ma X, Guo W (1983) {W-JS:} {A} modal logic of knowledge. In: Proceedings of the
  8th International Joint Conference on Artificial Intelligence. Karlsruhe,
  FRG, August 1983, pp 398--401

\bibitem[{McCarthy(1979)}]{McCarthy79}
McCarthy J (1979) First-{Order} theories of individual concepts and
  propositions. Machine Intelligence 9.:129--147

\bibitem[{Montgomery and Routley(1966)}]{MR66}
Montgomery H, Routley R (1966) Contingency and non-contingency bases for normal
  modal logics. Logique et Analyse 9:318--328

\bibitem[{Moore(1977)}]{Moore77}
Moore RC (1977) Reasoning about knowledge and action. In: Proceedings of
  IJCAI'77, pp 223--227

\bibitem[{Petrick and Bacchus(2004{\natexlab{a}})}]{Petrick-Bacchus:2004}
Petrick RPA, Bacchus F (2004{\natexlab{a}}) Extending the knowledge-based
  approach to planning with incomplete information and sensing. In: Zilberstein
  S, Koehler J, Koenig S (eds) Proceedings of ICAPS'04), {AAAI} Press, pp 2--11

\bibitem[{Petrick and Bacchus(2004{\natexlab{b}})}]{Petrick-Bacchus:2004b}
Petrick RPA, Bacchus F (2004{\natexlab{b}}) {PKS}: Knowledge-based planning
  with incomplete information and sensing. In: Proceedings of {ICAPS}'04

\bibitem[{Pizzi(2007)}]{Pizzi2007}
Pizzi C (2007) Necessity and relative contingency. Studia Logica 85(3):395--410

\bibitem[{Plaza(1989)}]{Plaza89:lopc}
Plaza JA (1989) Logics of public communications. In: Emrich ML, Pfeifer MS,
  Hadzikadic M, Ras ZW (eds) Proceedings of the 4th International Symposium on
  Methodologies for Intelligent Systems, pp 201--216

\bibitem[{Ryle(1949)}]{Ryle49}
Ryle G (1949) The Concept of Mind. Penguin

\bibitem[{Stanley(2011)}]{stanley2011know}
Stanley J (2011) Know how. Oxford University Press

\bibitem[{Stanley and Williamson(2001)}]{SW10}
Stanley J, Williamson T (2001) Knowing how. Journal of Philosophy 98:411--444

\bibitem[{Sturm et~al(2000)Sturm, Wolter, and Zakharyaschev}]{SturmWZ00}
Sturm H, Wolter F, Zakharyaschev M (2000) Monodic epistemic predicate logic.
  In: Proceedings of JELIA'00, pp 329--344

\bibitem[{V{\"{a}}{\"{a}}n{\"{a}}nen(2007)}]{Depbook}
V{\"{a}}{\"{a}}n{\"{a}}nen J (2007) Dependence Logic: A New Approach to
  Independence Friendly Logic. Cambridge University Press

\bibitem[{Van~Ditmarsch et~al(2011)Van~Ditmarsch, Herzig, and
  De~Lima}]{van2010situation}
Van~Ditmarsch H, Herzig A, De~Lima T (2011) From situation calculus to dynamic
  epistemic logic. Journal of Logic and Computation pp 179--204

\bibitem[{Von~Wright(1951)}]{Wright51}
Von~Wright GH (1951) An Essay in Modal Logic. North Holland, Amsterdam

\bibitem[{Wang(2015{\natexlab{a}})}]{Wang15:lori}
Wang Y (2015{\natexlab{a}}) A logic of knowing how. In: Proceedings of LORI-V,
  pp 392--405

\bibitem[{Wang(2015{\natexlab{b}})}]{Wang15:icla}
Wang Y (2015{\natexlab{b}}) Representing imperfect information of procedures
  with hyper models. In: Proceedings of ICLA'15, pp 218--231

\bibitem[{Wang(2017)}]{Wang17}
Wang Y (2017) A logic of goal-directed knowing how. Synthese Forthcoming

\bibitem[{Wang and Fan(2013)}]{WF13}
Wang Y, Fan J (2013) Knowing that, knowing what, and public communication:
  Public announcement logic with \texttt{Kv} operators. In: Proceedings of
  IJCAI'13, pp 1139--1146

\bibitem[{Wang and Fan(2014)}]{WF14}
Wang Y, Fan J (2014) Conditionally knowing what. In: Proceedings of AiML
  Vol.10, pp 569--587

\bibitem[{Wolter(2000)}]{Wolter00}
Wolter F (2000) First order common knowledge logics. Studia Logica
  65(2):249--271

\bibitem[{Xiong(2014)}]{Xiong14}
Xiong S (2014) Decidability of $\mathbf{ELKv^r}$. Bachelor's thesis, Peking
  University (in Chinese)

\bibitem[{Xu et~al(2016)Xu, Wang, and Studer}]{Xu16}
Xu C, Wang Y, Studer T (2016) A logic of knowing why, under submission

\bibitem[{Yu et~al(2016)Yu, Li, and Wang}]{YLW15}
Yu Q, Li Y, Wang Y (2016) A dynamic epistemic framework for conformant
  planning. In: Proceedings of TARK'15, EPTCS, forthcoming

\bibitem[{Zolin(1999)}]{Zolin99}
Zolin E (1999) Completeness and definability in the logic of noncontingency.
  Notre Dame Journal of Formal Logic 40(4):533--547

\bibitem[{Zolin(2001)}]{Zolin01}
Zolin E (2001) Infinitary expressibility of necessity in terms of contingency.
  In: Striegnitz K (ed) Proceedings of the Sixth ESSLLI Student Session, pp
  325--334

\end{thebibliography}
\bibliographystyle{spbasic}
\end{document}